\documentclass[11pt,a4paper]{article}

\usepackage{polyu_vclab}

\usepackage{lipsum}                
\usepackage{amsmath}
\usepackage{booktabs}
\usepackage{multirow}
\usepackage[numbers,sort&compress]{natbib}
\usepackage{floatflt}
\usepackage{algorithm}
\usepackage{algpseudocode}
\usepackage[capitalize]{cleveref}
\usepackage{graphicx}
\usepackage{tabularx}
\usepackage{wrapfig}
\usepackage{multirow}
\usepackage{enumitem}
\usepackage{amsmath}
\usepackage{booktabs}
\usepackage{xcolor}
\usepackage{pifont}
\usepackage{makecell}
\usepackage[table]{xcolor}         
\DeclareMathOperator*{\argmax}{arg\,max}
\Crefname{equation}{Eq.}{Eqs.} 
\crefname{equation}{Eq.}{Eqs.} 
\newcommand{\cmark}{\textcolor{green!70!black}{\ding{51}}}
\newcommand{\xmark}{\textcolor{red!70!black}{\ding{55}}}

\definecolor{forth}{RGB}{255,255,255}  
\definecolor{third}{RGB}{255,240,192} 
\definecolor{second}{RGB}{192,216,255}   
\definecolor{best}{RGB}{255,192,192}
\definecolor{lightgray}{RGB}{180, 180, 180}
\setEyebrow{PolyU VCLab\,\textbullet\,Preprint 2026}

\setReportTag{Visual Computing Lab\,\textperiodcentered\,The Hong Kong Polytechnic University}

\papertitle{\raisebox{-0.3em}{\includegraphics[height=1.2em]{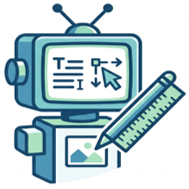}}Text-Vision Co-Instructed Image Editing}

\paperauthors{%
  Chenxi Xie\affilmark{1,2}\quad
  Yuhui Wu\affilmark{1,2}\quad
  Qiaosi Yi\affilmark{1,2}\quad
  Lei Zhang\correspondmark\affilmark{1,2}%
}

\paperaffil{%
  \affilmark{1}\,The Hong Kong Polytechnic University\quad
  \affilmark{2}\,OPPO Research Institute
}

\papernotes{%
  \correspondmark\,Corresponding author (\href{mailto:cslzhang@comp.polyu.edu.hk}{cslzhang@comp.polyu.edu.hk}).%
}

\paperbadges{%
  \vclabbadgesolid{Project Page}{https://xiechenxi99.github.io/TVEdit/}\;%
  \vclabbadgesolid{Code}{https://github.com/PolyU-VCLab/TVEdit}\;%
  \vclabbadgesolid{Hugging Face}{https://huggingface.co/VCLab-PolyU/TVEdit/tree/main}\;%
}

\begin{document}

\maketitleVCLab

\begin{vclabAbstract}
\noindent\textbf{Abstract.}\;
Existing image editing methods can be generally categorized into textual instruction-based and visual prompt-based ones. 
Textual instructions are semantically expressive, but are limited by the coarse granularity of spatial control of the editing results. In contrast, visual prompts such as drag and point can  provide precise spatial guidance, but are limited by the inherent ambiguity in semantic intent. 
To unify the strength of textual and visual prompts, we present \textbf{Text-Vision Co-Instructed Image Editing}, which jointly models textual instructions as semantic intent and sparse visual instructions as spatial guidance, aiming to achieve precise and intent-faithful image manipulation.
To this end, we first construct a textual-visual instruction paired dataset with more than 23K samples derived from dynamic videos, enabling aligned supervision for cross-modal instruction.
We then propose \textbf{TV-Edit}, a \textbf{T}extual-\textbf{V}isual instruction unified \textbf{Edit}ing framework to contextualize drag or point-based visual instructions with image-text semantics and lift them into semantic-aware control representations for pretrained editing backbones. 
By integrating semantic intent and spatial constraints, TV-Edit leads to more precise spatial control, less instruction ambiguity, and stronger structural consistency than text-only or drag-based alternatives. 
Finally, we establish \textbf{TV-Edit-Bench}, a deliberately designed benchmark to evaluate semantic faithfulness, spatial alignment, and visual consistency with ground-truth references and controlled textual–visual variations for reliable assessment.
Our experiments across multiple editing backbones demonstrate that TV-Edit consistently yields more precise and intent-faithful edits, significantly outperforming state-of-the-art instruction-based and drag-based baselines.
\end{vclabAbstract}





\section{Introduction}
\begin{figure}[t] 
    \centering
    \includegraphics[width=0.95\linewidth]{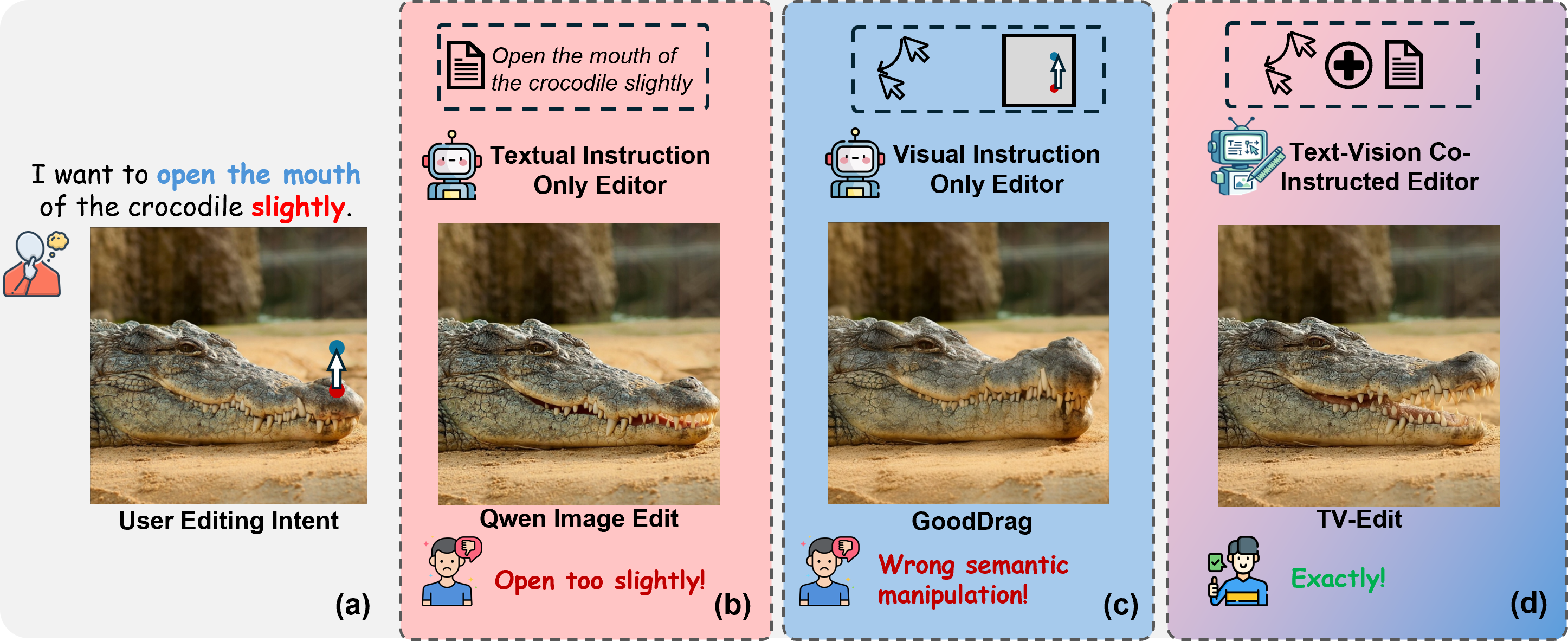}
    \caption{Comparison among different editing paradigms in terms of inputs and results. (a) User editing intent. (b) Textual instruction-based editing. \cite{wu2025qwen} (c) Visual prompt-based editing \cite{zhang2024gooddrag}. (d) Our proposed text-vision co-instructed editing (TV-Editing).}
    
    \label{fig:intro}
    \vspace{-5mm}
\end{figure}

Benefiting from the rapid development of text-to-image (T2I) generation models \cite{song2020ddim,liu2023flow,rombach2022high} and multi-modal large language models (MLLMs) \cite{bai2025qwen3,gpt4o,liu2024llava15,nanobanana2025}, image editing has made significant progress in recent years \cite{kulikov2024flowedit,xie2025dnaedit,kawar2023imagic,brooks2023instructpix2pix}. In particular, instruction-based image editing  \cite{labs2025flux,liu2025step1x,wu2025qwen} has achieved remarkable success, allowing users to specify the intent of editing through expression in natural language. These models are highly effective at manipulating semantic attributes such as color, material, and category with strong visual fidelity. However, natural language remains limited in specifying editing effects that involve spatial control or object actions, such as location, pose, shape, and motion changes. As illustrated in \cref{fig:intro} (b), Qwen-Image-Edit \cite{wu2025qwen} fails to quantify the user's intent of ``slightly'', leading to edits that deviate from the user's expectation.

On the other hand, drag-based editing \cite{pan2023drag,hou2024easydrag,zhang2024gooddrag,xia2025draglora,mou2023dragondiffusion,jiang2024clipdrag} offers an effective alternative by allowing precise spatial control. As a representative form of sparse visual prompt-based editing, it allows users to specify target displacements through drag points, achieving fine-grained control over object layout. However, unlike textual instruction-based editors, these methods are predominantly designed for point-only input, which can specify local motion, but provides little semantic grounding.  As shown in \cref{fig:intro}(c), when a user draws an upward arrow intending to open the crocodile's upper jaw, drag-based methods instead produce an unintended deformation of the jaw region. Although it satisfies the geometric constraint, it fails to capture the intended semantic action. 

Therefore, these two paradigms share a common limitation: a single modality is insufficient to fully convey user intent. Textual instructions are well suited to specifying \emph{what} a semantic transformation should occur, yet limited in determining \emph{where} and \emph{how} it should be implemented. Sparse visual instructions, in contrast, explicitly constrain local motion and geometry, but limited in specifying the intended semantic transformation. The representational limitation of textual and visual instructions makes it difficult for each of them alone to faithfully realize complex editing goals.

Motivated by the above observations, we propose \textbf{Text-Vision Co-instructed Image Editing}, a new task where textual instructions specify semantic intent, and sparse visual prompts impose spatial constraints. Rather than relying exclusively on textual instructions or visual prompts, we treat them as complementary signals to jointly specify the desired editing effects. Our goal is to achieve more precise and intent-faithful image manipulation by reducing the ambiguity inherent in single-modality intent expression.
To this end, we introduce \textbf{TV-Edit}, a \textbf{T}extual-\textbf{V}isual instruction unified \textbf{Edit}ing framework. We first construct a large-scale textual-visual instruction paired dataset from video data, obtaining more than 23K quadruplets of $\{$\textit{source image}, \textit{target image}, \textit{point trajectory}, \textit{text prompt}$\}$. Leveraging the temporal continuity of videos, we derive sparse geometric cues through point tracking and optical flow, and pair them with content-aware semantic edit descriptions generated by MLLM \cite{bai2025qwen3}. This paired dataset explicitly binds textual semantic intent and sparse visual constraints to the same target edit, enabling unified instruction learning. We then present a decoupled Content-Aware Spatial Controller. Rather than directly injecting sparse point trajectories as geometry-only control signals \cite{shi2024lightningdrag, zhang2025framepainter}, the controller integrates image content, textual conditions, and geometric cues to produce dense control features for pretrained editing backbones. Such a design makes TV-Edit plug-and-play with modern instruction-based foundation editors, equipping them with semantically grounded spatial control. As illustrated in \cref{fig:intro}(d), our TV-Editing model produces edits that are semantically faithful, spatially aligned, and ultimately more consistent with user intent.

To systematically evaluate existing methods and our TV-Edit on this new task, we introduce \textbf{TV-Edit-Bench}. {It contains 120 curated evaluation pairs drawn from multiple sources, including real videos, image-to-video generated videos, and image pairs synthesized by advanced editing models, providing rich and diverse editing intents.} Each sample of TV-Edit-Bench provides aligned textual and visual instructions together with ground-truth editing targets, enabling reliable assessment of semantic faithfulness, spatial alignment, and visual consistency. Furthermore, the benchmark features with two sub-tasks for geometry disambiguation and semantic disambiguation, explicitly testing a method's ability to follow both types of constraints.

We apply TV-Edit to popular editing foundation models, including Qwen-Image-Edit \cite{wu2025qwen} and FLUX.1 Kontext \cite{labs2025flux}.
Experimental results show that TV-Edit consistently outperforms existing state-of-the-art instruction-based and drag-based competitors, producing edits that are more faithful to the semantic intent and the spatial constraints specified by users.

\section{Related Work}

\paragraph{Textual Instruction-based Editing}
has evolved from early prompt-based methods \cite{xie2025dnaedit, ju2023direct, mokady2023null} to instruction-based models. Early approaches rely on pre-trained text-to-image (T2I) models via inversion and regeneration, requiring carefully aligned source and target prompts. As a more effective alternative, instruction-based editing replaces cumbersome prompt pairs with direct commands. Pioneering works like InstructP2P \cite{brooks2023instructpix2pix, yu2025anyedit,ye2025imgedit} train models on paired datasets via diffusion or flow matching, bypassing per-instance inversion. Recently, foundation models such as FLUX.1 Kontext \cite{labs2025flux} and Qwen-Image-Edit \cite{wu2025qwen} have substantially improved instruction following and visual fidelity. However, these models still struggle with edits involving complex actions and continuous motions. To tackle this, ByteMorph \cite{chang2025bytemorph} introduces a dataset and baseline for non-rigid motions, while MotionEdit \cite{wan2025motionedit} proposes a motion-centric benchmark and post-training framework. Despite these advances, natural language inherently lacks the precision to describe fine-grained dynamics (e.g., exact motion magnitudes or trajectories), often leading to under-specified spatial realizations. This fundamental limitation motivates our formulation of text-vision co-instructed editing.

\noindent\textbf{Visual Prompt-based Editing}
enables intuitive image manipulation through sparse visual inputs such as points, strokes, and sketches \cite{zhang2025framepainter,pan2023drag}. Among these paradigms, drag-based editing has become one of the most representative forms. 
Existing diffusion-based drag methods can be broadly divided into two categories. Optimization-based methods \cite{hou2024easydrag,xia2025draglora,zhang2024gooddrag,mou2023dragondiffusion} rely on inversion and test-time latent optimization, whereas training-based methods \cite{shi2024lightningdrag,zhang2025framepainter} learn motion priors from videos and therefore enable more efficient inference. However, both categories primarily enforce geometric constraints without explicitly modeling semantic intent, making edits ambiguous. 
Recent works have begun to incorporate semantics, but their progress remains limited. 
CLIP-Drag \cite{jiang2024clipdrag} introduces global text guidance, but its coarse CLIP-based \cite{radford2021clip} gradients fail to align reliably with fine-grained trajectories. Drag-Flow \cite{zhou2025dragflow} leverages DiT priors \cite{flux.1-dev-weights} and MLLM assistance, yet requires dedicated mask inputs and only resolves task-level ambiguity.
Moreover, these methods still operate within an optimization-based drag-editing framework, making them sensitive to per-instance hyperparameters. More fundamentally, they treat semantics as an auxiliary extension to drag-based editing and are evaluated mainly on drag-oriented benchmarks that emphasize geometric accuracy. In contrast, we formalize text-vision co-instructed editing as a unified task and establish a dedicated framework, dataset, and benchmark for jointly evaluating semantic alignment and spatial controllability, enabling stable, mask-free, and efficient editing.

\section{Text-Vision Co-Instructed Editing}

\subsection{Problem Formulation}
Conventional image editing is typically formulated as a conditional mapping problem, i.e., $\hat{\mathcal{I}}_{tgt} = f_\theta(\mathcal{I}_{src}, c)$, where $\mathcal{I}_{src}$ is the source image and condition $c$ is either a textual instruction or a sparse visual prompt. Such a formulation is inherently under-specified for complex edits, where a single modality-based prompt cannot fully express user intent.
In this work, we propose \textbf{Text-Vision Co-Instructed Image Editing} (TV-Edit), where the desired edit is specified jointly by textual and visual instructions. Given a source image $\mathcal{I}_{src} \in \mathbb{R}^{H \times W \times 3}$, a textual instruction $\mathbf{t}_{\mathrm{txt}}$ describing the intended semantic transformation, and a sparse visual prompt $\mathcal{P}_{\mathrm{vis}}=\{(\mathbf{p}^{\mathrm{src}}_k,\mathbf{p}^{\mathrm{tgt}}_k)\}_{k=1}^{K}$, where $\mathbf{p}^{\mathrm{src}}_k,\mathbf{p}^{\mathrm{tgt}}_k \in \mathbb{R}^2$ denote the source and target coordinates of the $k$-th control point, the goal is to learn a model to generate an edited image as follows:
\begin{equation}
\hat{\mathcal{I}}_{tgt} = f_\theta(\mathcal{I}_{src}, \mathbf{t}_{\mathrm{txt}}, \mathcal{P}_{\mathrm{vis}}).
\end{equation}

The edited image $\hat{\mathcal{I}}_{tgt}$ is expected to satisfy three requirements: (i) semantic faithfulness to $\mathbf{t}_{\mathrm{txt}}$; (ii) spatial alignment with \(\mathcal{P}_{\mathrm{vis}}\) while maintaining locally coherent semantic transformations; and (iii) global coherence with $\mathcal{I}_{src}$ beyond the primary edited regions.

\subsection{Text-Vision Paired Data Construction}

\begin{figure}[t]
    \centering
    \includegraphics[width=\linewidth]{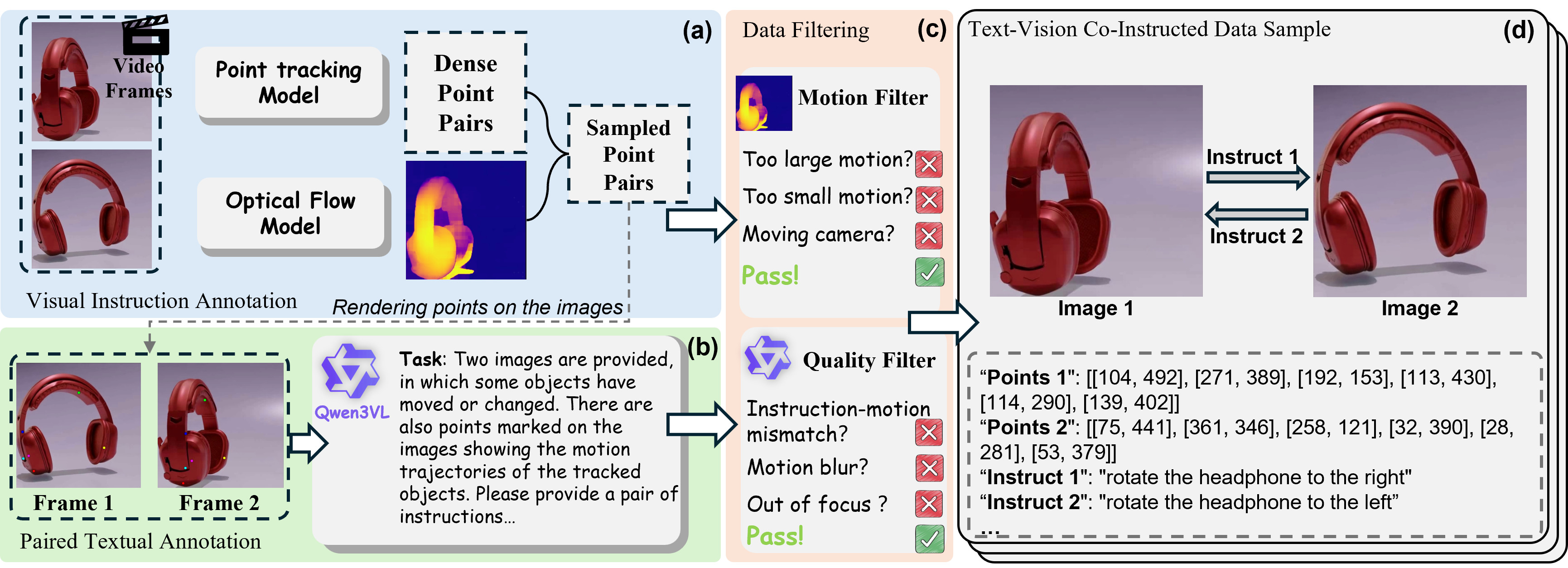}
    \caption{\textbf{TV-Edit-23K data construction pipeline.} Given two video frames, we first perform (a) visual instruction annotation to obtain sparse point pairs. We then conduct (b) paired textual annotation on the image pair with rendered points to get motion-focused instructions aligned with the marked points. After (c)  data filtering, we obtain (d) the text-vision co-instructed data.}
    \label{fig:pipeline}
\vspace{-4mm}
\end{figure}

Since existing editing datasets \cite{yu2025anyedit,shi2024lightningdrag,ye2025imgedit} are designed mainly for single-modality supervision, we thus construct a textual-visual instruction paired training dataset to provide cooperative supervision to learn our TV-Edit model. Considering that videos naturally contain rich and continuous motion changes and deformations, we collect video data from open source video datasets \cite{xue2025ultravideo,ravi2024sam,ditto} to build our dataset. 
Specifically, we segment full videos into clips with varying temporal strides and extract the initial and final frames. 
From the pairs of initial and final frames, we construct training quadruplets through three stages: visual instruction annotation, paired textual annotation, and data filtering, which are illustrated in \cref{fig:pipeline} and detailed as follows.

\noindent \textbf{Visual Instruction Annotation.} As shown in \cref{fig:pipeline} (a), we use the SEA-RAFT \cite{wang2024sea} and the Co-Tracker-V3 \cite{karaev2025cotracker3} to obtain the optical flow magnitude map and the dense grid point trajectories, respectively. As in \cite{shi2024lightningdrag,zhang2025framepainter}, the normalized flow-magnitude map is utilized as a spatial sampling weight to filter dense points,  
which retains significant movements while preserving small displacements, yielding sparse point pairs that accurately track cross-frame motion.

\noindent \textbf{Paired Textual Annotation.} 
Directly prompting an MLLM with raw image pairs often yields descriptions misaligned with the intended motion, due to complex video dynamics and inherent MLLM hallucinations.
Inspired by recent works \cite{shtedritski2023does, cai2024vip}, we introduce a visual prompting strategy. As shown in \cref{fig:pipeline} (b), we render the points filtered in the first stage onto the image pairs using different colors and design corresponding prompts to guide the MLLM. This ensures that the MLLM focuses exclusively on the desired dynamics, thereby achieving aligned text-vision annotations.

\noindent \textbf{Data Filtering.} To ensure the quality of generated samples, we perform motion, alignment, and visual quality filtering. Because image editing focuses on static scenes, we address the common issue of global video dynamics by thresholding optical flow maps. Specifically, we discard samples where the frame, particularly the boundary regions, shows significant optical flow. This ensures that we only keep images with static backgrounds. To verify text annotations, we introduce a closed-loop generate-then-verify paradigm \cite{yin2024woodpecker} to minimize MLLM hallucinations. Lastly, we perform a basic screening for motion blur and overall image quality to finalize the dataset.

Ultimately, we collect 23K high-quality sample groups, which build our dataset, namely \textbf{TV-Edit-23K}. As illustrated in \cref{fig:pipeline}(d), each raw sample group, denoted as \((I_1, I_2, p_1, p_2, T_1, T_2)\), can be decoupled into two bidirectional editing pairs for supervision. Specifically, they are formulated as the forward mapping \((I_1, p_1, p_2, T_1) \rightarrow I_2\) and the backward mapping \((I_2, p_2, p_1, T_2) \rightarrow I_1\). The constructed TV-Edit-23K dataset encompasses not only diverse scenes but also rich semantic motion transformations and varying motion magnitudes. It provides high-quality, text-vision paired supervision for the training of our TV-Edit model. More details can be found in \textbf{Appendix A}.

\subsection{TV-Edit Model Training}

\begin{figure}[t]
    \centering
    \includegraphics[width=\linewidth]{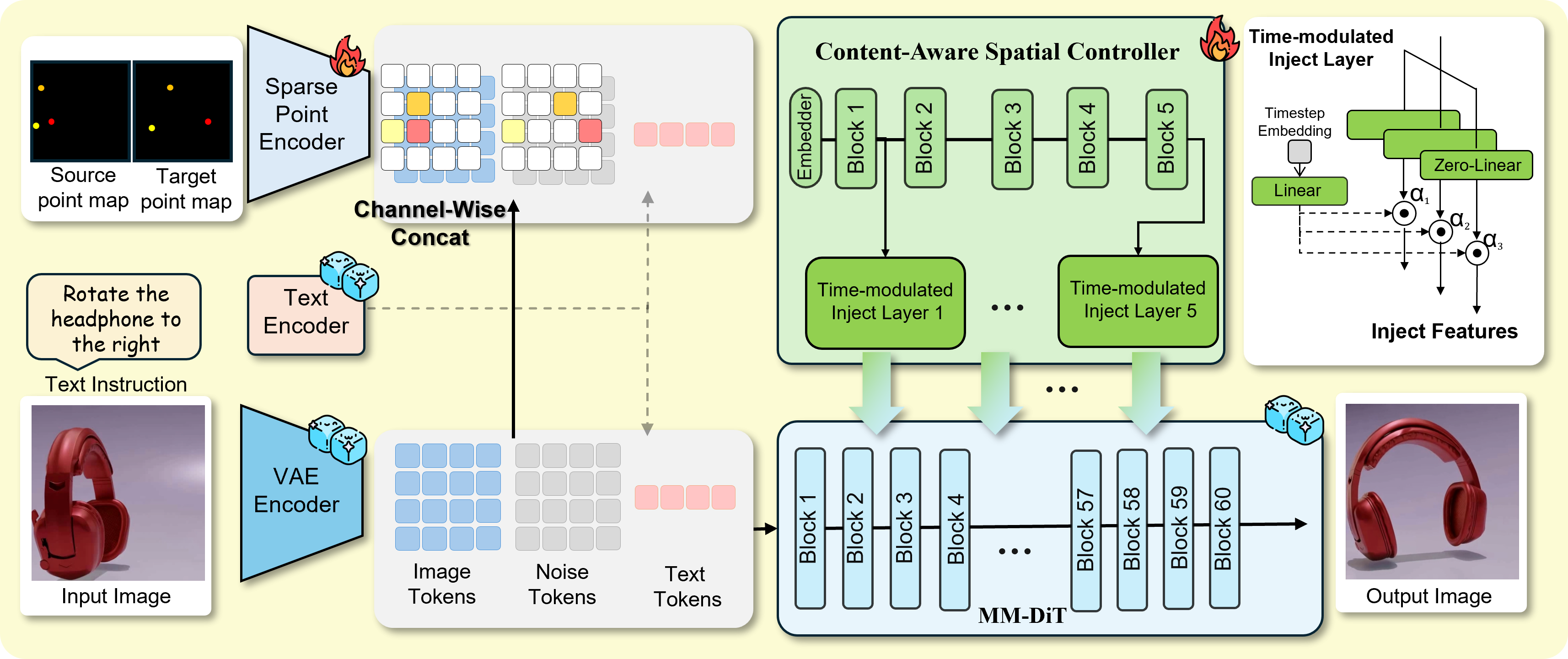}
    \caption{\textbf{The architecture of our TV-Edit.} TV-Edit consists of a main branch and a control branch. In the main branch, image, text, and noised latent tokens are processed by the editing backbone. In the control branch, source and target point maps are encoded and concatenated with image and noised latent tokens, grounding sparse trajectories with image content. A lightweight Content-Aware Spatial Controller then performs a early fusion and produces control features for the backbone.}
    \label{fig:main}
\end{figure}

\textbf{Model Architecture}.
Recent textual instruction-based editing methods \cite{wu2025qwen,labs2025flux,liu2025step1x} have demonstrated strong semantic modeling abilities to understand user instructions; however, their editing process lacks fine-grained spatial controllability. Therefore, rather than training a model from scratch or fine-tuning a base text-to-image (T2I) model, we introduce a control branch to inject spatial intent, thereby guiding the editing foundation model to produce geometrically grounded results.

As shown in \cref{fig:main}, our TV-Edit model consists of two branches: a main editing branch, which includes a VAE encoder, a text encoder and a Multi-Modal Diffusion Transformer (MM-DiT), and a control branch, which includes a sparse point encoder and a content-aware controller.
Given a source image $\mathcal{I}_{src} \in \mathbb{R}^{H \times W \times 3}$, a textual instruction $\mathbf{t}_{\mathrm{txt}}$, and a sparse visual prompt $\mathcal{P}_{\mathrm{vis}}=\{(\mathbf{p}^{\mathrm{src}}_k,\mathbf{p}^{\mathrm{tgt}}_k)\}_{k=1}^{K}$, where $\mathbf{p}^{\mathrm{src}}_k,\mathbf{p}^{\mathrm{tgt}}_k \in \mathbb{R}^{2}$ denote the source and target coordinates of the $k$-th control point, the source image is first encoded into image latent tokens $ \mathbf{Z}_{\mathrm{img}} \in \mathbb{R}^{\frac{H}{8}\times\frac{W}{8}\times C} $ by the VAE encoder, and the textual instruction is encoded into text tokens $ \mathbf{X}_{\mathrm{txt}} $ by the text encoder. In the main editing branch, $ \mathbf{Z}_{\mathrm{img}} $, $ \mathbf{X}_{\mathrm{txt}} $, and the noised latent tokens $ \mathbf{Z}_{\mathrm{noise}} $ are fed into the MM-DiT backbone for denoising.

In the control branch, the sparse visual prompt $\mathcal{P}_{\mathrm{vis}}$ is first rendered as two spatial maps, $\mathbf{M}_{\mathrm{src}}, \mathbf{M}_{\mathrm{tgt}} \in \mathbb{R}^{H \times W \times 1}$. To preserve the correspondence between source and target points, the pixel value at each control point is set to its trajectory index, while all the other locations are set to zero:
\begin{equation}
\mathbf{M}_{c}[x,y] =
\begin{cases}
k, & \text{if } (x,y)=\mathbf{p}^{c}_k,\\
0, & \text{otherwise},
\end{cases}
\qquad c \in \{\mathrm{src}, \mathrm{tgt}\}.
\end{equation}
The two maps are then processed by a sparse point encoder composed of lightweight convolutional layers. The encoder produces spatially compressed point embeddings $\mathbf{E}_{\mathrm{src}}, \mathbf{E}_{\mathrm{tgt}} \in \mathbb{R}^{\frac{H}{8} \times \frac{W}{8} \times C}$, which have the same spatial resolution as the image latent.
The source and target point embeddings are concatenated with the image latent and noised latent, respectively, followed by linear projection:
\begin{equation}
\hat{\mathbf{Z}}_{\mathrm{img}} = \phi\big([\mathbf{Z}_{\mathrm{img}} ; \mathbf{E}_{\mathrm{src}}]\big), \qquad
\hat{\mathbf{Z}}_{\mathrm{noise}} = \phi\big([\mathbf{Z}_{\mathrm{noise}} ; \mathbf{E}_{\mathrm{tgt}}]\big),
\end{equation}
where $[\cdot ; \cdot]$ denotes channel-wise concatenation and $\phi(\cdot)$ is a linear projection for channel compression.

The above operations ground the semantic-agnostic point correspondences in the source image content. As a result, the controller can interpret spatial prompts in a content-dependent manner rather than taking them as geometry-only signals. The controller then produces several residual control features and injects them into the MM-DiT backbone in a ControlNet-like manner to guide the spatial composition of the edited image.
The plug-and-play design of our TV-Edit model allows it to be seamlessly integrated into multiple popular editing foundation models. By optimizing only a small number of trainable parameters, TV-Edit can leverage the strong priors of the pretrained backbone while maintaining high training efficiency.

\noindent \textbf{Lightweight Content-aware Spatial Controller}.
Directly injecting sparse spatial cues into the main diffusion backbone is suboptimal, as the foundation model struggles to align rigid, semantic-agnostic spatial relations with highly semantic visual features. To address this, we design a content-aware spatial controller. By performing early fusion over spatial cues, image content, and textual conditions, it produces content-aware guidance features to accurately guide the editing process.

Inspired by ControlNet \cite{zhang2023adding}, we adopt MM-DiT blocks aligned with the backbone as the controller. 
To keep the controller lightweight, we reduce its complexity in two ways. First, we halve the hidden dimension, which reduces the parameter count by nearly 75\%. Second, we restrict the number of network blocks to $N$, which is significantly less than the number of blocks, denoted by  $N_{\mathrm{main}}$, in the main backbone.
However, this aggressive compression may reduce the expressiveness of the controller outputs, weakening their ability to regulate the backbone effectively. To mitigate this issue, we introduce a \emph{time-modulated inject layer}. This design adaptively regulates the injection intensity at different semantic depths via block-wise scaling, while dynamically aligning the control influence with the temporal evolution of the diffusion process. For the output $\mathbf{H}_n$ of the $n$-th controller block, we attach three block-specific linear projection heads $\phi_{n,i}(\cdot)$, where $i \in \{1,2,3\}$. In parallel, we predict three timestep-dependent modulation coefficients from the timestep embedding $\mathbf{e}_t$:
\begin{equation}
\alpha_{n,i} = g_{n,i}(\mathbf{e}_t), 
\qquad i \in \{1,2,3\},\; n \in \{1,\dots,N\},
\end{equation}
where $g_{n,i}(\cdot)$ is a learnable mapping from the timestep embedding to a scalar coefficient. Each projected feature is then scaled by its corresponding coefficient:
\begin{equation}
\mathbf{F}_{n,i} = \alpha_{n,i}\,\phi_{n,i}(\mathbf{H}_n),
\qquad i \in \{1,2,3\},\; n \in \{1,\dots,N\},
\end{equation}
where $\mathbf{F}_{n,i}$ denotes the resulting modulated control feature used for injection into the main backbone.

Finally, there are a total of $3N$ modulated inject features. To match the number of blocks \(N_{\mathrm{main}}\) in the main backbone, the output features from each controller block are repeated \(\frac{N_{\mathrm{main}}}{3N}\) times before being injected into the corresponding layers of MM-DiT. In practice, each backbone block receives a residual control feature, enabling dense guidance throughout the denoising process.

\noindent \textbf{Training Strategy}.
As shown in \cref{fig:main}, during training we freeze the main branch and optimize only the sparse point encoder and the controller from scratch. Our task prioritizes spatial layout, which is predominantly determined in the high-noise regime. We therefore adopt the \(\mathbf{Z}_0\)-prediction objective, as it equates to a \(t^2\)-weighted velocity loss that intentionally assigns larger weights to large \(t\):
\begin{equation}
    \mathcal{L}_{\mathrm{fm}}
    =
    \mathbb{E}_{t,\mathbf{Z}_0,\mathbf{Z}_1}
    \left[
    \left\|
    \hat{\mathbf{Z}}_0 - \mathbf{Z}_0
    \right\|_2^2
    \right],
    \qquad
    \hat{\mathbf{Z}}_0 = \mathbf{Z}_t - t \cdot v_\theta(\mathbf{Z}_t, \mathbf{Z}_{\mathrm{img}}, \mathbf{X}_{\mathrm{txt}}, \mathbf{E}_{\mathrm{src}}, \mathbf{E}_{\mathrm{tgt}}, t),
\end{equation}
where $\mathbf{Z}_0$ is the clean latent of the target image, $\mathbf{Z}_1$ is the sampled Gaussian noise, and $\mathbf{Z}_t = (1-t)\mathbf{Z}_0 + t\mathbf{Z}_1$. 
Furthermore, to explicitly reinforce this structural focus, we replace uniform sampling with a Beta distribution \(t \sim \mathrm{Beta}(\alpha,\beta)\) biased toward larger timesteps during early training. This bias is gradually relaxed as training progresses. In practice, this strategy improves both convergence speed and controllability. More details are provided in the \textbf{Appendix B}.

\section{Experiments}
\subsection{Experimental Setup}

\textbf{Implementation Details}.
We implement TV-Edit on both Qwen-Image-Edit \cite{wu2025qwen} and FLUX.1 Kontext \cite{labs2025flux}, demonstrating its generality across different editing backbones. TV-Edit is trained for 80K iterations with AdamW \cite{loshchilov2017decoupled} as the optimizer, using a learning rate of $1 e^{-4}$. The effective batch size is set to 64 via gradient accumulation. The Beta noise sampling schedule is annealed from $\mathrm{Beta}(20,2)$ to $\mathrm{Beta}(5,2)$ over the first 40K iterations and kept fixed afterwards. We use a number of $N=5$ controller blocks, which provide a good trade-off between controllability and efficiency. Training is conducted on 16 NVIDIA A800 GPUs.

\noindent \textbf{Compared Methods.} TV-Edit is a plug-and-play extension to instruction-based foundation editors, aiming to improve their spatial control capability. Note that TV-Edit does not sacrifice the foundation models' original semantic editing ability, such as adding/deleting objects or changing their attributes.  We thus compare TV-Edit against two groups of related baselines: drag-based methods and advanced instruction-based editing models. For drag-based methods, we compare with a few representative and state-of-the-art approaches, such as optimization-based methods DragDiffusion \cite{mou2023dragondiffusion} and GoodDrag \cite{zhang2024gooddrag}, and training-based method LightningDrag \cite{shi2024lightningdrag}. For instruction-based models, we compare against state-of-the-art editing models, including LongCat-Image-Edit \cite{team2025longcat}, FLUX-Kontext \cite{batifol2025flux} and Qwen-Image-Edit \cite{wu2025qwen}. In addition, we include MotionEdit \cite{wan2025motionedit}, which is specifically designed for motion scenarios, and the powerful closed-source commercial model NanoBananaPro \cite{nanobanana2025}.

\noindent \textbf{Evaluation Benchmarks.} Since there is no existing benchmark available for text-vision co-instructed editing models, we carefully constructed such a benchmark, namely \textbf{TV-Edit-Bench}. The details are provided in Sec. \ref{TV-edit-bench}. 
On the other hand, since TV-Edit is designed for spatially related editing, we further evaluate it on established drag-based benchmarks to verify the generalization of its spatial control ability. Quantitative and qualitative results are provided in the \textbf{Appendix F}.


\subsection{TV-Edit-Bench}
\label{TV-edit-bench}

\begin{figure}
    \centering
    \includegraphics[width=\linewidth]{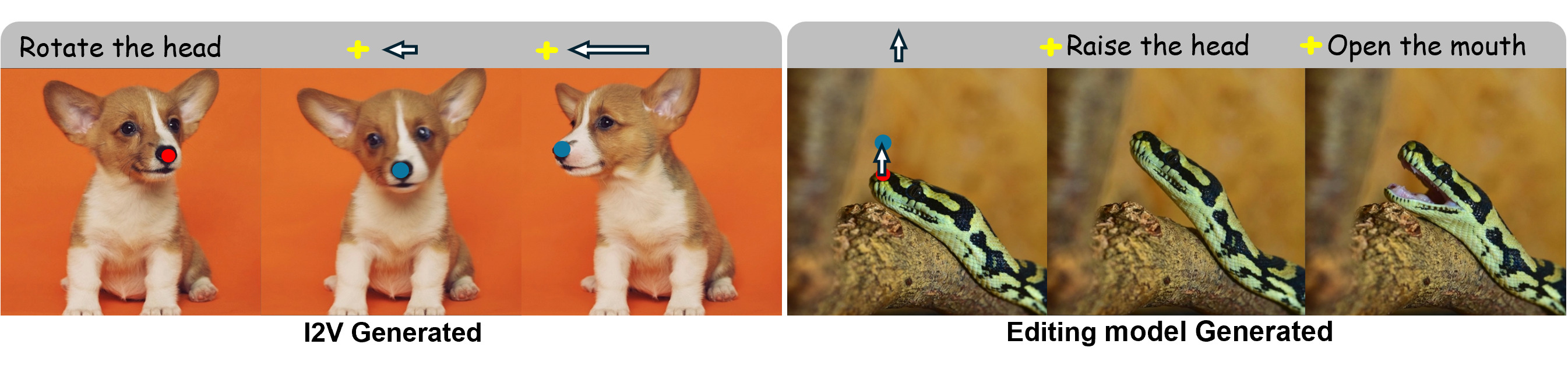}
    \vspace{-6mm}
    \caption{Examples of the two sub-tasks in TV-Edit-Bench. Left: same textual instruction with different motion magnitudes. Right: similar visual prompts with different semantic instructions.}
    \label{fig:benchdata}
    \vspace{-5mm}
\end{figure}

We construct \textbf{TV-Edit-Bench}, a benchmark for evaluating text-vision co-instructed editing with paired textual and visual instructions. For the limit of space, here we provide the basic information of its \textit{Data Curation} process and \textit{Evaluation Protocol}. More details are provided in the \textbf{Appendix C}.

\noindent \textbf{Data Curation.}  We collect candidate image pairs from three sources, each serving a different purpose. First, to capture realistic scene dynamics and natural motion patterns, we sample frame pairs from real videos \cite{xue2025ultravideo}. Then, to explicitly evaluate the model's capability in fine-grained multimodal control, we further design two specific sub-tasks. The first sub-task tests motion magnitude control under fixed semantic intents. We use Wan2.2 \cite{wan2025wan} to generate examples where the source image undergoes identical actions but with varying spatial extents (see \cref{fig:benchdata}, left). The second sub-task tests semantic disambiguation under similar visual prompts. We use NanoBanana Pro \cite{nanobanana2025} to create instances with similar  trajectories but distinct semantic transformations (see \cref{fig:benchdata}, right). Note that these pairs are strictly filtered and manually checked before used as benchmark cases. 

With the collected image pairs, we follow a pipeline similar to \cref{fig:pipeline} to annotate sparse trajectories and textual instructions with a strict manual screening to retain only high-quality samples with consistent images, trajectories, and instructions. In total, TV-Edit-Bench contains 120 carefully curated samples. Each sample provides paired textual and visual prompts together with a ground-truth edited target. We further include auxiliary annotations such as masks and descriptions to facilitate inference and evaluation for different editing methods. 

\noindent \textbf{Evaluation Protocol.}
We evaluate TV-Editing from three complementary perspectives: image fidelity, geometric accuracy, and semantic faithfulness.

\textit{\# Image fidelity.}
We assess image fidelity with LPIPS \cite{lpips} against the ground-truth target image. To reduce the influence of pixel misalignment, we further introduce two DINOv3-based \cite{simeoni2025dinov3} metrics , namely \(\mathrm{DS}_{\mathrm{global}}^{\mathrm{tgt}}\) and \(\mathrm{DS}_{\mathrm{local}}^{\mathrm{tgt}}\), which measure global and local visual consistency, respectively.

\textit{\# Geometric accuracy.}
We evaluate geometric accuracy by measuring the displacement error between matched points in the edited image and the target points. Specifically, we report a sparse matching distance \(\mathrm{MD}_s\) and a dense matching distance \(\mathrm{MD}_d\), both normalized by the image size, where matching is performed using high-resolution DINOv3 features within restricted local regions around the handle and target points to avoid false matches.

\textit{\# Semantic faithfulness.}
We use an MLLM-based evaluator to assess whether the edited result is semantically faithful to the intended edit. Following recent MLLM-based evaluation practices \cite{he2025contextdrag,fu2024dreamsim}, we report two scores, concept preservation (CP) and prompt following (PF).


\definecolor{rowblue}{HTML}{EAF4FF}
\definecolor{roworange}{HTML}{F6EDE2}
\definecolor{rowred}{HTML}{FDECEC}
\definecolor{rowpurple}{HTML}{F3ECFF}

\begin{table}[t]
    \centering
    \caption{Quantitative comparison on TV-Edit-Bench. M, P, T, I denote mask, prompt, trajectory and instruction respectively. The best and second-best results are highlighted in \textbf{bold} and \underline{underlined}.}
    \label{tab:drag_results}
    \resizebox{0.95\linewidth}{!}{%
    \begin{tabular}{l c | c c c | c c | c c}
        \toprule
        \multirow{2}{*}{\textbf{Method}} 
        & \multirow{2}{*}{\textbf{Input}} 
        & \multicolumn{3}{c|}{\textbf{Image Fidelity}} 
        & \multicolumn{2}{c|}{\textbf{Geometric Accuracy}} 
        & \multicolumn{2}{c}{\textbf{MLLM Score}} \\
        & & DS$_{\mathrm{global}}^{\mathrm{tgt}}$\,$\uparrow$ & DS $_{\mathrm{local}}^{\mathrm{tgt}}$\,$\uparrow$ & LPIPS$^{\mathrm{tgt}}$\,$\downarrow$ 
        & MD$_\mathrm{s}$\,$\downarrow$ & MD$_\mathrm{d}$\,$\downarrow$ 
        & PF\,$\uparrow$ & CP\,$\uparrow$ \\
        \midrule
        \rowcolor{rowblue}
        DragDiffusion \cite{mou2023dragondiffusion}   & M+P+T    & .8290 & .9288 & .2490 & .1044 & .1048 & 0.72 & 0.98 \\
        \rowcolor{rowblue}
        GoodDrag \cite{zhang2024gooddrag}       & M+P+T  & .8724 & .9305 & .2234 & .0630 & .0648 & 0.75 & 0.96 \\
        \rowcolor{rowblue}
        DragLoRA \cite{xia2025draglora}       & M+P+T  & .8444 & .9207 & .2380 & .0638 & .0671 & 0.77 & 0.94 \\
        \rowcolor{rowblue}
        \rowcolor{rowblue}
        \rowcolor{rowblue}
        CLIP-Drag \cite{jiang2024clipdrag}       & M+P+T   & .8741 & .9301   & .2312    & .1383    & .1363    & 0.71    & 0.95    \\
        \rowcolor{rowblue}
        GeoDrag \cite{pu2025dragging}        & M+P+T  & .8710 & .9059 & .2904 & .0775 & .0776 & 0.79 & 0.94 \\
        \rowcolor{rowblue}
        \rowcolor{rowblue}
        LightningDrag \cite{shi2024lightningdrag}  & M+T    & .8254 & .9155 & .2633 & .0649 & .0689 & 0.79 & 0.93 \\
        \midrule
        \rowcolor{rowred}
        FLUX-Kontext \cite{labs2025flux}     & I   & .8740 & .9302   & .2728    & .1626    & .1613    & 0.82    & 0.98    \\
        \rowcolor{rowred}
        Qwen-Image-Edit \cite{wu2025qwen}         & I   & .8575 & .9191   & .3123    & .1379    & .1380    & 0.86    & 0.97    \\
        
        \rowcolor{rowred}
        LongCat-Image-Edit \cite{team2025longcat} &I &.8550 &.9185 &.3391 &.1247 &.1229 & \textbf{0.93} & \underline{0.99} \\
        \rowcolor{rowred}
         MotionEdit  \cite{wan2025motionedit} &I &.8459 &.9166 &.2943 &.1228 &.1173 & \underline{0.92} &0.97 \\
        \rowcolor{rowred}
        NanoBananaPro \cite{nanobanana2025} &I &\underline{.9096} &.9432 &.2072 &.1201 &.1195 &{0.89} &\textbf{1.00} \\
        \midrule
        \rowcolor{rowpurple}
        TV-Edit-Kontext  & I+T  & \textbf{.9134}  & \textbf{.9514}  & \underline{.1696}    & \underline{.0484}    & \underline{.0508}    & 0.86 & \underline{0.99} \\
        \rowcolor{rowpurple}
        TV-Edit-Qwen     & I+T  & \textbf{.9134} & \underline{.9490} &\textbf{.1672} & \textbf{.0421} & \textbf{.0462} & \textbf{0.93} & \textbf{1.00} \\
        \bottomrule
    \end{tabular}%
    }
    \vspace{-4mm}
\end{table}

\subsection{Main Results}

\textbf{Quantitative Results.} In \cref{tab:drag_results}, we present a quantitative comparison of drag-based methods, instruction-based methods, and our TV-Edit on the TV-Edit-Bench. One can see that Drag-based methods achieve strong geometric accuracy, but their image fidelity is often compromised. Furthermore, despite precise trajectory tracking, their PF scores remain low, indicating poor semantic execution. Specifically, while the leading method GoodDrag\cite{zhang2024gooddrag} achieves an excellent $\mathrm{MD}_d$ of 0.0648, its PF score is only 0.75. This reveals that drag-based methods can reliably control \textit{where} to edit, but struggle to determine \textit{what} semantic action to perform.

Instruction-based models exhibit the opposite trend. They achieve strong image fidelity and semantic faithfulness, as reflected by NanoBanana Pro, which reaches 0.9432 on \(\mathrm{DS}_{\mathrm{global}}^{\mathrm{tgt}}\), nearly 1.0 on CP, and 0.89 on PF.  However, relying entirely on textual instructions without explicit spatial guidance makes their geometric changes highly unpredictable. Consequently, their \(\mathrm{MD}_d\) consistently exceeds 0.10. In other words, these models can determine \textit{what} semantic edit to perform, but cannot precisely control \textit{where} it is spatially realized.

By contrast, our TV-Edit significantly bridges this gap, achieving high geometric accuracy and semantic faithfulness simultaneously. TV-Edit-Qwen reduces \(\mathrm{MD}_d\) to 0.0462, achieving a 28.7\% improvement over the best drag-based method and demonstrating excellent spatial precision. Beyond spatial control, it elevates the PF score from 0.86 to 0.93 compared to its base model Qwen-Image-Edit, even surpassing the closed-source model NanoBanana Pro. This indicates that the visual input not only provides explicit geometric control, but also works collaboratively with textual guidance to improve semantic execution. Notably, TV-Edit is mask-free but still preserves strong image fidelity, leading to stable and faithful edits.

\begin{figure}[t]
    \centering
    \includegraphics[width=\linewidth]{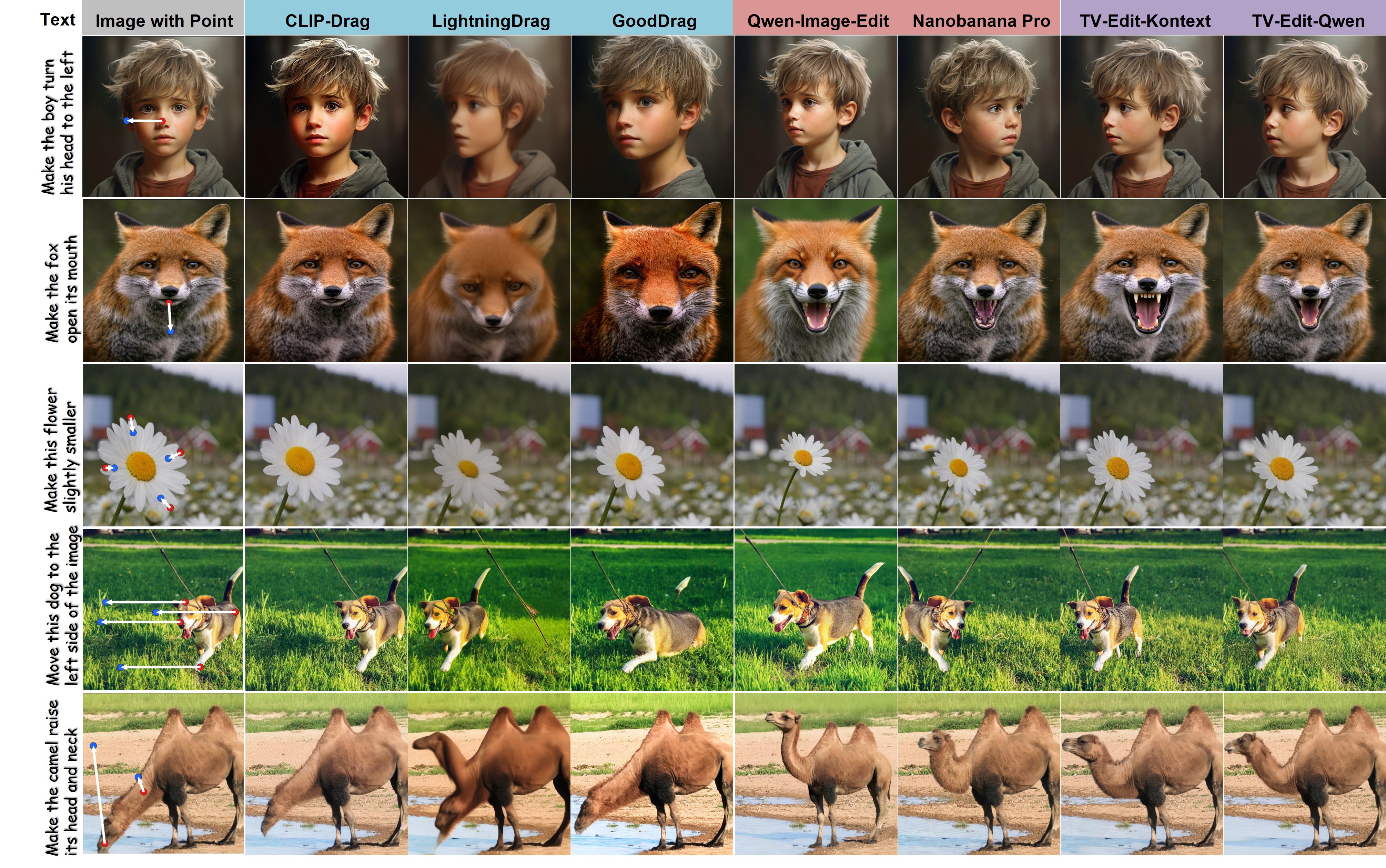}
    \caption{Qualitative results of competing methods on TV-Edit-Bench.}
    \label{fig:main_vcomp}
    \vspace{-4mm}
\end{figure}

\noindent \textbf{Qualitative Results}. We first present visual comparisons with state-of-the-art methods to highlight TV-Edit's superiority in accurate semantic editing and precise spatial control, then demonstrate its flexibility on fine-grained multimodal  control by the two sub-tasks.

In \cref{fig:main_vcomp}, we present several representative spatially related editing cases, including rotation, non-rigid action, scale change, and translation. Drag-based methods can produce reasonable results when there is little semantic ambiguity of the editing trajectory. For example, in the 1st row, most drag-based baselines correctly rotate the boy's head. However, once the visual prompt admits multiple semantic realizations, these methods often fail to capture the intended action. In the 2nd row, instead of opening the fox's mouth, drag-based methods distort the facial region or produce implausible edits. In addition, the optimization-based point-tracking methods struggle with large motions. In the 4th row, GoodDrag attempts to deform the dog rather than translate it, while LightningDrag moves the dog but leaves the leash behind, resulting in obvious artifacts.
On the other hand, instruction-based editing methods Qwen-Image-Edit and Nanobanana Pro correctly capture the semantic intent described by the text. However, they often fail to match the desired motion extent or direction. For instance, NanoBanana Pro rotates the boy's head in the wrong direction in the 1st row and shrinks the flower too much in the 3rd row. In the 4th row, it also alters the dog's pose and layout, deviating from the original intent.
By contrast, both versions of our TV-Edit produce edits that are semantically correct, spatially precise, and visually coherent. They can more accurately control the magnitude of the action while preserving the image fidelity. In the 4th row, even without explicit control over the leash, TV-Edit moves it consistently with the dog, yielding faithful results.

\begin{figure}[t]
    \centering
    \includegraphics[width=0.95\linewidth]{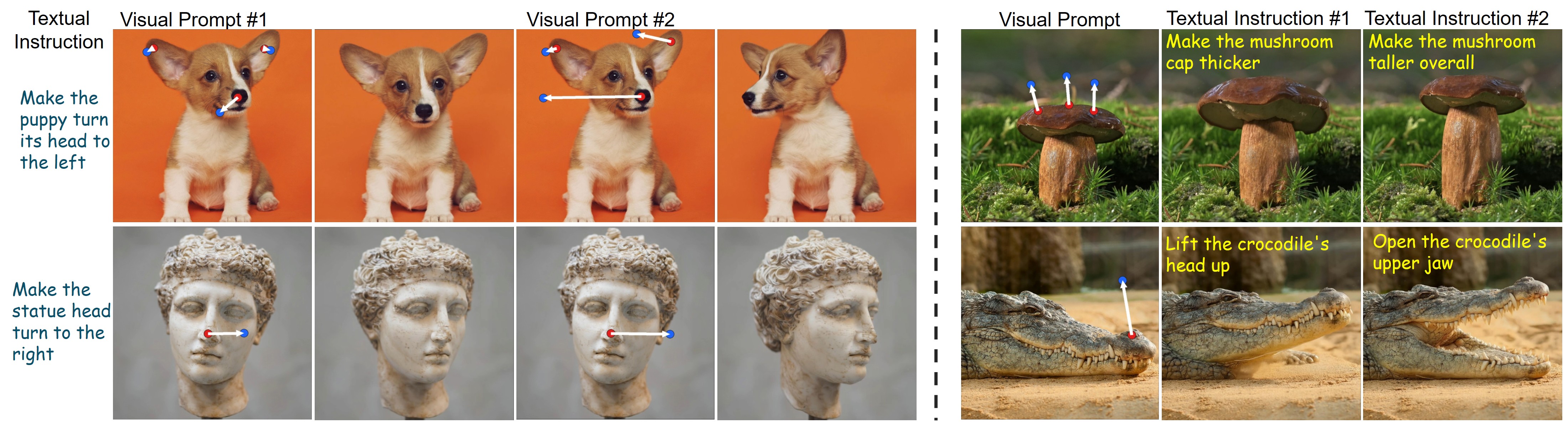}
    \caption{Editing results of TV-Edit on the two sub-tasks of fine-grained control. Left: motion magnitude variation task. Right: semantic variation task.}
    \label{fig:main_sub_task}
    \vspace{-4mm}
\end{figure}

\cref{fig:main_sub_task} demonstrates TV-Edit's flexibility on the two sub-tasks of fine-grained control.  On the left, it accurately controls motion magnitudes under a fixed textual instruction (\textit{e.g.}, rotating the dog's head to different degrees). On the right, when visual prompts are ambiguous, it adapts to different texts to realize distinct actions (\textit{e.g.}, lifting the head vs. opening the mouth for the same trajectory). These results show that TV-Edit can accurately realize user intent by leveraging textual and visual prompts.

In \textbf{Appendix D}, we provide more qualitative comparisons. 
\subsection{{Ablation Studies on TV-Edit}}

We conduct ablation studies to validate the architectural design of TV-Edit and our choice of hyperparameters. All experiments are conducted using the Qwen-Image-Edit-based version of TV-Edit. Additional ablations on multimodal input strategies at inference time, and isolating the contribution of dataset from that of the architecture and training paradigm, are provided in \textbf{Appendix E}.

\textbf{Ablation of Architecture}. In \cref{tab:ablation_arch}, we present an ablation study on our proposed architecture. We establish a baseline that includes only a sparse point encoder and linear layers that inject features into the editing backbone. As shown in the first row, simply injecting sparse geometric conditions is insufficient for effective spatial control, resulting in a suboptimal geometric accuracy where $\text{MD}_{\text{d}}$ only reaches 0.1355. 
As can be seen in the second row, incorporating the content-aware spatial controller yields substantial improvements in both semantic consistency and geometric accuracy. This indicates that rather than relying on sparse trajectories, processing the fused trajectory and image content through a transformer can yield highly effective control features for backbone injection. Subsequently, in the third row, we attempt to expand the output features of each block using multiple linear injectors before feeding them into the main branch. As observed in the third row, this modification only brings marginal gains. Despite the addition of extra linear layers intended to increase feature diversity, the network struggles to learn distinct representations without explicit guidance.
As indicated in the fourth row, we introduce a time-modulated scaling mechanism, which substantially enhances the model's control capabilities. During the denoising process, it dynamically adjusts the influence of the control branch features on different backbone blocks over timesteps. This dynamic modulation effectively balances spatial control and detail generation, ultimately leading to editing results with high visual consistency.

\begin{table}[t]
  \centering
  \caption{Ablation study on architecture.}
  \label{tab:ablation_arch}
  \begin{tabular}{lcccc}
    \toprule
    \textbf{Setting} & $\text{DS}_{\text{global}}^{\text{tar}}\!\uparrow$ & $\text{DS}_{\text{local}}^{\text{tar}}\!\uparrow$ & $\text{MD}_{\text{d}}\!\downarrow$ & $\text{PF}\!\uparrow$ \\
    \midrule
    baseline   & .8595          &  .9185        & .1355         & 0.89  \\
    + content aware spatial controller   &   .9024    &    .9430       &.0505          &0.87  \\
    + multiple linear injectors  & .9015 &.9443 & .0501 &0.91  \\
    + time-modulated injectors  & \textbf{.9134}      &  \textbf{.9490}        & \textbf{.0462}       & \textbf{0.93} \\
    \bottomrule
  \end{tabular}
\end{table}

\begin{table}[t]
  \small
  \setlength{\tabcolsep}{3pt}
  \begin{minipage}[t]{0.48\textwidth}
    \centering
    \caption{Ablation on noise sampling strategies.}
    \label{tab:ablation_noise}
    \begin{tabular}{l| c c c c}
      \toprule
      \textbf{Noise Dist} & DS$_{\mathrm{local}}^{\mathrm{tar}}$\,$\uparrow$ & DS$_{\mathrm{global}}^{\mathrm{tar}}$\,$\uparrow$ & MD$_\mathrm{d}$\,$\downarrow$ & PF\,$\uparrow$ \\
      \midrule
      $\mathrm{Beta}(1,1)$  & .9324 & .8857 & .0568 & 0.90 \\
      $\mathrm{Beta}(3,2)$  & .9379 & .8966 & .0510 & 0.90 \\
      $\mathrm{Beta}(5,2)$  & \textbf{.9490} & \textbf{.9134} & \textbf{.0462} & \textbf{0.93} \\
      $\mathrm{Beta}(10,2)$ & .9335 & .8861 & .0604 & 0.88 \\
      \bottomrule
    \end{tabular}
  \end{minipage}%
  \hfill
  \begin{minipage}[t]{0.48\textwidth}
    \centering
    \caption{Ablation on the number of blocks.}
    \label{tab:ablation_b}
    \begin{tabular}{l| c c c c}
      \toprule
      \textbf{Block Num} & DS$_{\mathrm{local}}^{\mathrm{tar}}$\,$\uparrow$ & MD$_\mathrm{d}$\,$\downarrow$ & PF\,$\uparrow$ & \# Para. \\
      \midrule
       1&.9369  &{.0481}  &0.88  &167M  \\
       3&.9401  &.0549  &0.91  &337M  \\
       5&.\textbf{9490}  &\textbf{.0462}  &\textbf{0.93}  &506M  \\
       15&.9400  &{.0468}  &0.90  &1.40B  \\
      \bottomrule
    \end{tabular}
    
  \end{minipage}
  \vspace{-4mm}
\end{table}

\textbf{Ablation of Different Timestep Sampling Strategies}. \cref{tab:ablation_noise} shows the effect of different timestep sampling strategies during training. Since TV-Editing mainly involves low-frequency structural changes, placing more emphasis on larger timesteps generally improves geometric accuracy. This trend is reflected by the clear gain under \(\mathrm{Beta}(5,2)\). However, when the sampling distribution becomes overly concentrated on high-noise steps, both geometric accuracy and prompt-following performance deteriorate. Although high-noise stages are crucial for structural control, later timesteps also make important contributions to the editing process.

\textbf{Ablation of the Number of Controller Blocks.} \cref{tab:ablation_b} shows the effect of the number of controller blocks. Even with only one block, the controller already achieves strong performance, reaching an \(\mathrm{MD}_d\) of 0.0481 with only 167M trainable parameters, which demonstrates the efficiency of our architecture design. As the number of blocks increases, the prompt-following score further improves and reaches 0.93 at \(N=5\), indicating stronger semantic control. However, when \(N\) is increased to 15, the parameter count rises to 1.4B and the prompt-following score drops under the same training iteration. Considering both performance and efficiency, we use \(N=5\) in our TV-Edit.



\section{Conclusion}
We introduced \textbf{Text-Vision Co-Instructed Image Editing}, a new image editing task that jointly leverages textual instructions and sparse visual prompts to reduce the ambiguity of single-modality control. To tackle this task, we proposed \textbf{TV-Edit}, a plug-and-play framework with a decoupled Content-Aware Spatial Controller, and constructed a text-vision paired training dataset to learn unified semantic and spatial control. In addition, we curated \textbf{TV-Edit-Bench}, which contains carefully designed cases and a comprehensive evaluation protocol. Extensive experiments on TV-Edit-Bench revealed the limitations of both visual prompt-only and instruction-only editing methods. By contrast, TV-Edit achieved stronger semantic faithfulness, spatial controllability, and visual consistency, demonstrating its superiority over existing methods for robust image editing.

\noindent \textbf{Limitations}. Our TV-Edit has some limitations. First, it is built upon large-scale editing foundation models so that its inference speed prevents from real-time interactive editing. Second, it works well for 2D operations but remains limited for complex 3D manipulations such as out-of-plane rotations.


{\small
\bibliography{ref}
}


\newpage
\appendix

\setcounter{figure}{0}
\setcounter{table}{0}

\renewcommand{\thefigure}{A\arabic{figure}}
\renewcommand{\thetable}{A\arabic{table}}
\setcounter{equation}{0}
\renewcommand{\theequation}{A\arabic{equation}}

\begin{center}
    \LARGE \textbf{Appendix}
\end{center}

In this appendix, we provide the following materials:
\begin{enumerate}[label=\Alph*]
    \item  More details of the TV-Edit-23K dataset (referring to Sec. 3.2 in the main paper).
    \item More analysis of training strategy (referring to Sec. 3.3 in the main paper).
    \item  More details of TV-Edit-Bench, including the dataset construction and evaluation protocol (referring to Sec. 4.2 in the main paper).
    \item  More editing results of TV-Edit and more visual comparisons on TV-Edit-Bench (referring to Sec. 4.3 in the main paper).
    \item  More ablation studies on TV-Edit (referring to Sec. 4.4 in the main paper).
    \item Quantitative and qualitative comparison with drag-based methods on Drag-Bench. 
    \item User study.
    \item Failure cases and analysis.
    \item  Potential social impact.
\end{enumerate}

\section{More Details of TV-Edit-23K Dataset}

\subsection{Detailed Prompts for Paired Textual Annotation}
\begin{figure}[h]
    \centering
    \includegraphics[width=\linewidth]{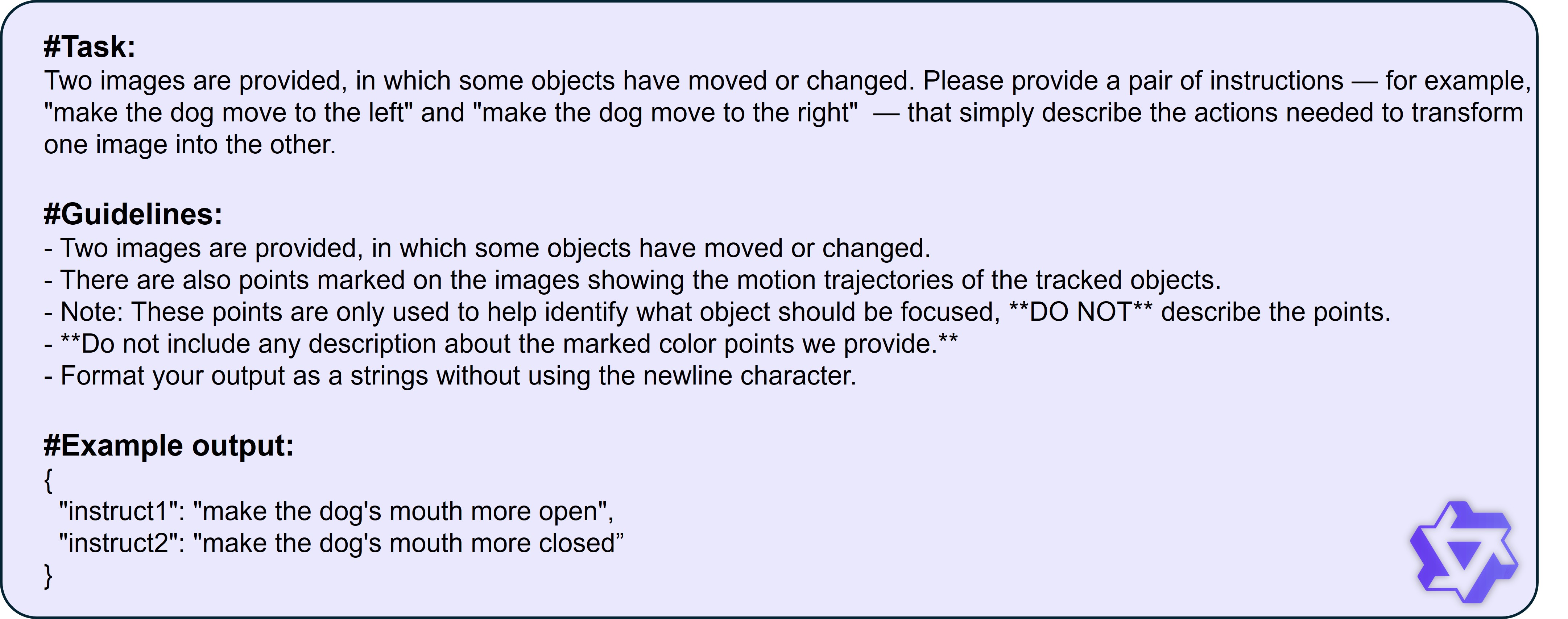}
    \caption{Detailed prompts for paired textual annotation in TV-Edit-23K data construction pipeline.}
    \label{fig:app_datapipe}
\end{figure}

In the paired textual annotation stage, we instruct Qwen-3-VL \cite{bai2025qwen3} to provide the action that can transform one image to the other. Detailed prompts for this transformation are shown in \cref{fig:app_datapipe}.

\subsection{Samples in TV-Edit-23K Dataset}
\begin{figure}[H]
    \centering
    \includegraphics[width=1\linewidth]{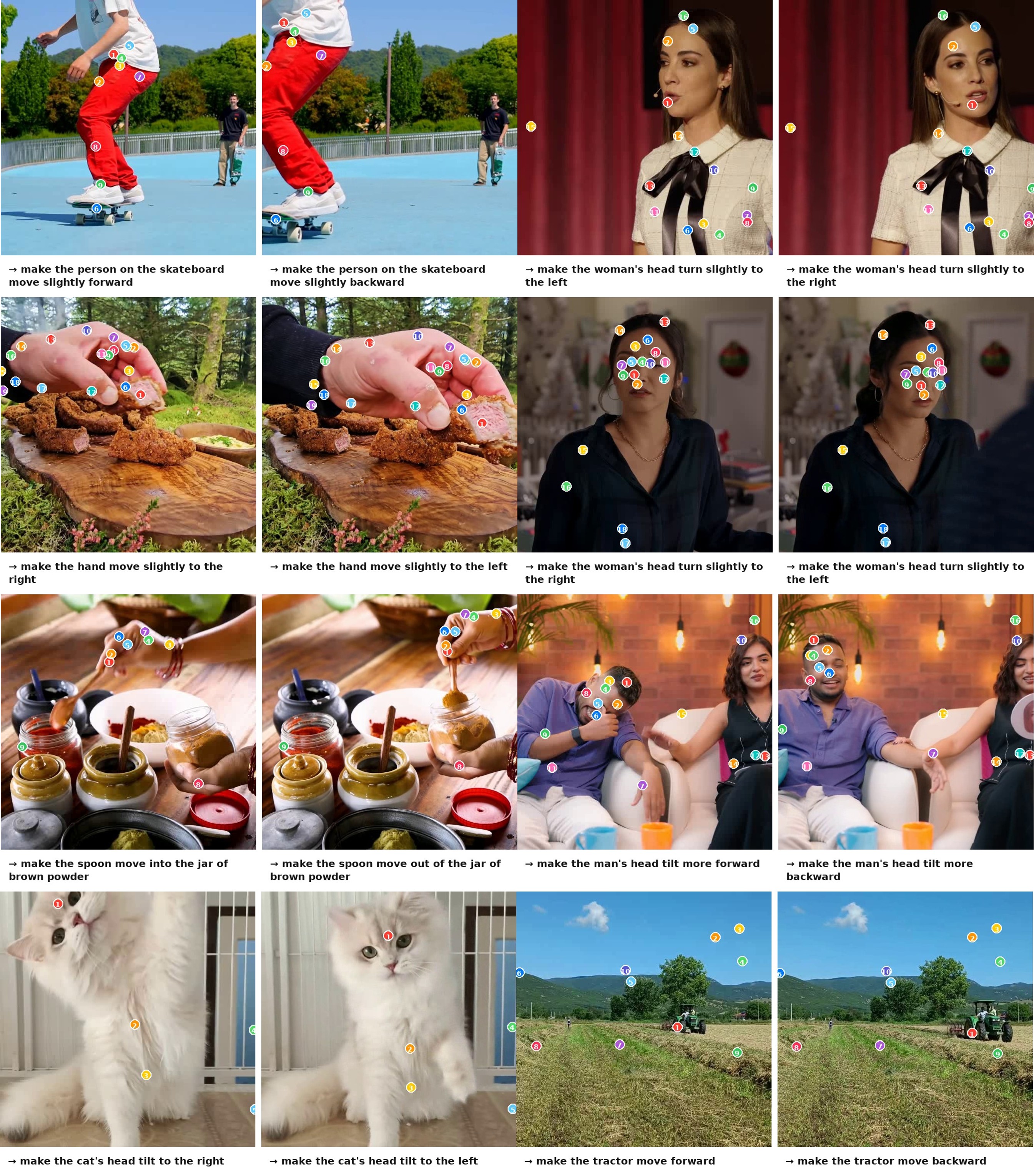}
    \caption{Samples in TV-Edit-23K Dataset. The text below the image represents the instructions to transform the image into another image.}
    \label{fig:app_train}
\end{figure}

In \cref{fig:app_train}, we show some training samples in TV-Edit-23K. We see that our data construction pipeline generates training data with dense and accurate point pairs, enabling the model to learn the corresponding geometric relationships. Furthermore, it can be seen that MLLM provides accurate semantic transformation instructions which match the motion of labeled point pairs. 

\section{More Analysis of TV-Edit Training Strategy}

Since TV-Edit aims to achieve precise spatial control and geometric accuracy guided by sparse points, prioritizing the optimization of the global spatial layout is crucial. In flow matching models, the generative process at large timesteps (the high-noise regime, where \(t \to 1\)) is primarily responsible for establishing this global spatial layout and low-frequency structures. Conversely, small timesteps (the low-noise regime, where \(t \to 0\)) focus on high-frequency texture details. Therefore, the fundamental motivation behind our training strategy is to force the network to focus its learning capacity on the high-noise regime. We achieve this through two complementary approaches: an implicit loss weighting mechanism and an explicit time-step sampling strategy.

\begin{figure}[H]
    \centering
    \includegraphics[width=0.4\linewidth]{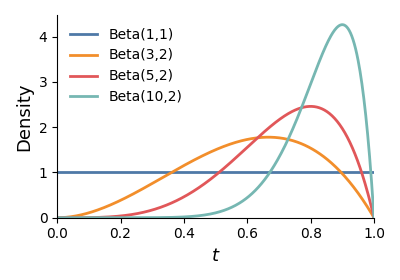}
    \caption{Illustration of different Beta distribution.}
    \label{fig:appen_beta}
\end{figure}

\textbf{V-supervision flow matching loss.} 
In the standard Rectified Flow, the forward process constructs a linear interpolation between the clean data latent \(\mathbf{Z}_0\) and the sampled Gaussian noise \(\mathbf{Z}_1\). The intermediate state \(\mathbf{Z}_t\) at timestep \(t \in [0, 1]\) is defined as:
\begin{equation}
    \mathbf{Z}_t = (1 - t)\mathbf{Z}_0 + t\mathbf{Z}_1.
\end{equation}
The ground-truth velocity field  that drives \(\mathbf{Z}_0\) to \(\mathbf{Z}_1\) is simply the derivative of \(\mathbf{Z}_t\) with respect to \(t\):
\begin{equation}
    v_t = \frac{\mathrm{d}\mathbf{Z}_t}{\mathrm{d}t} = \mathbf{Z}_1 - \mathbf{Z}_0.
\end{equation}
Typically, the model \(v_\theta\) is trained to predict this constant velocity using the standard \(v\)-prediction objective, which uniformly weights all timesteps:
\begin{equation}
    \mathcal{L}_{v} = \mathbb{E}_{t, \mathbf{Z}_0, \mathbf{Z}_1} \left[ \left\| v_\theta(\mathbf{Z}_t) - (\mathbf{Z}_1 - \mathbf{Z}_0) \right\|_2^2 \right].
\end{equation}

\textbf{Derivation of the \(\mathbf{Z}_0\)-supervision flow matching loss.}
Instead of directly supervise the velocity, TV-Edit adopts a \(\mathbf{Z}_0\)-supervision loss. Based on the forward process definition, we can express the ground-truth \(\mathbf{Z}_0\) in terms of \(\mathbf{Z}_t\) and the ground-truth velocity \(v_t\). By substituting \(\mathbf{Z}_1 = \mathbf{Z}_0 + v_t\) into the \(\mathbf{Z}_t\) equation, we get:
\begin{equation}
    \mathbf{Z}_t = (1 - t)\mathbf{Z}_0 + t(\mathbf{Z}_0 + v_t) = \mathbf{Z}_0 + t v_t.
\end{equation}
Thus, the ground-truth data latent can be written as:
\begin{equation}
    \mathbf{Z}_0 = \mathbf{Z}_t - t v_t.
\end{equation}
During training, our model predicts the velocity \(v_\theta(\mathbf{Z}_t, \dots)\), which is then used to estimate the clean latent \(\hat{\mathbf{Z}}_0\):
\begin{equation}
    \hat{\mathbf{Z}}_0 = \mathbf{Z}_t - t \cdot v_\theta(\mathbf{Z}_t)
\end{equation}
Our training objective \(\mathcal{L}_{\mathrm{fm}}\) minimizes the Mean Squared Error (MSE) between the estimated \(\hat{\mathbf{Z}}_0\) and the ground-truth \(\mathbf{Z}_0\). By substituting the expressions for \(\hat{\mathbf{Z}}_0\) and \(\mathbf{Z}_0\), we can reveal its relationship with the standard v-supervision flow matching loss:

\begin{align}
    \mathcal{L}_{\mathrm{fm}} &= \mathbb{E}_{t, \mathbf{Z}_0, \mathbf{Z}_1} \left[ \left\| \hat{\mathbf{Z}}_0 - \mathbf{Z}_0 \right\|_2^2 \right] \nonumber \\
    &= \mathbb{E}_{t, \mathbf{Z}_0, \mathbf{Z}_1} \left[ \left\| (\mathbf{Z}_t - t \cdot v_\theta(\mathbf{Z}_t)) - (\mathbf{Z}_t - t \cdot v_t) \right\|_2^2 \right] \nonumber\\
    &= \mathbb{E}_{t, \mathbf{Z}_0, \mathbf{Z}_1} \left[ \left\| - t \cdot v_\theta(\mathbf{Z}_t) + t \cdot v_t \right\|_2^2 \right] \nonumber\\
    &= \mathbb{E}_{t, \mathbf{Z}_0, \mathbf{Z}_1} \left[ t^2 \left\| v_\theta(\mathbf{Z}_t) - (\mathbf{Z}_1 - \mathbf{Z}_0) \right\|_2^2 \right].
\end{align}

This demonstrates that our \(\mathbf{Z}_0\)-supervision is mathematically equivalent to the \(v\)-supervision loss scaled by a coefficient \(t^2\). This implicit \(t^2\) mechanism intentionally assigns significantly larger penalty weights to errors made at large \(t\). The effectiveness of this loss weighting strategy has also been corroborated by recent studies in controllable generation~\cite{sangare2026improving}.

\textbf{Timestep sampling strategy.}
While the \(\mathbf{Z}_0\)-supervision implicitly weights the loss, we further explicitly bias the training distribution to sample more large timesteps. As illustrated in \cref{fig:appen_beta}, we plot the probability density functions of the Beta distribution under various \(\alpha\) and \(\beta\) hyperparameters. For \(\text{Beta}(10, 2)\), the distribution is heavily concentrated around \(0.9\), which corresponds to extremely high noise levels where spatial layout is determined. Conversely, the \(\text{Beta}(3, 2)\) distribution is only slightly skewed towards larger timesteps, and empirical training under this setting is inferior to \(\text{Beta}(5, 2)\). As evidenced by the ablation study on sampling distributions presented in this appendix, the \(\text{Beta}(5, 2)\) strategy yields the optimal performance for TV-Edit, as it primarily focuses on the high-noise regime while maintaining adequate coverage of low-noise timesteps.



\section{More Details of TV-Edit-Bench}
\begin{figure}[t]
    \centering
    \includegraphics[width=\linewidth]{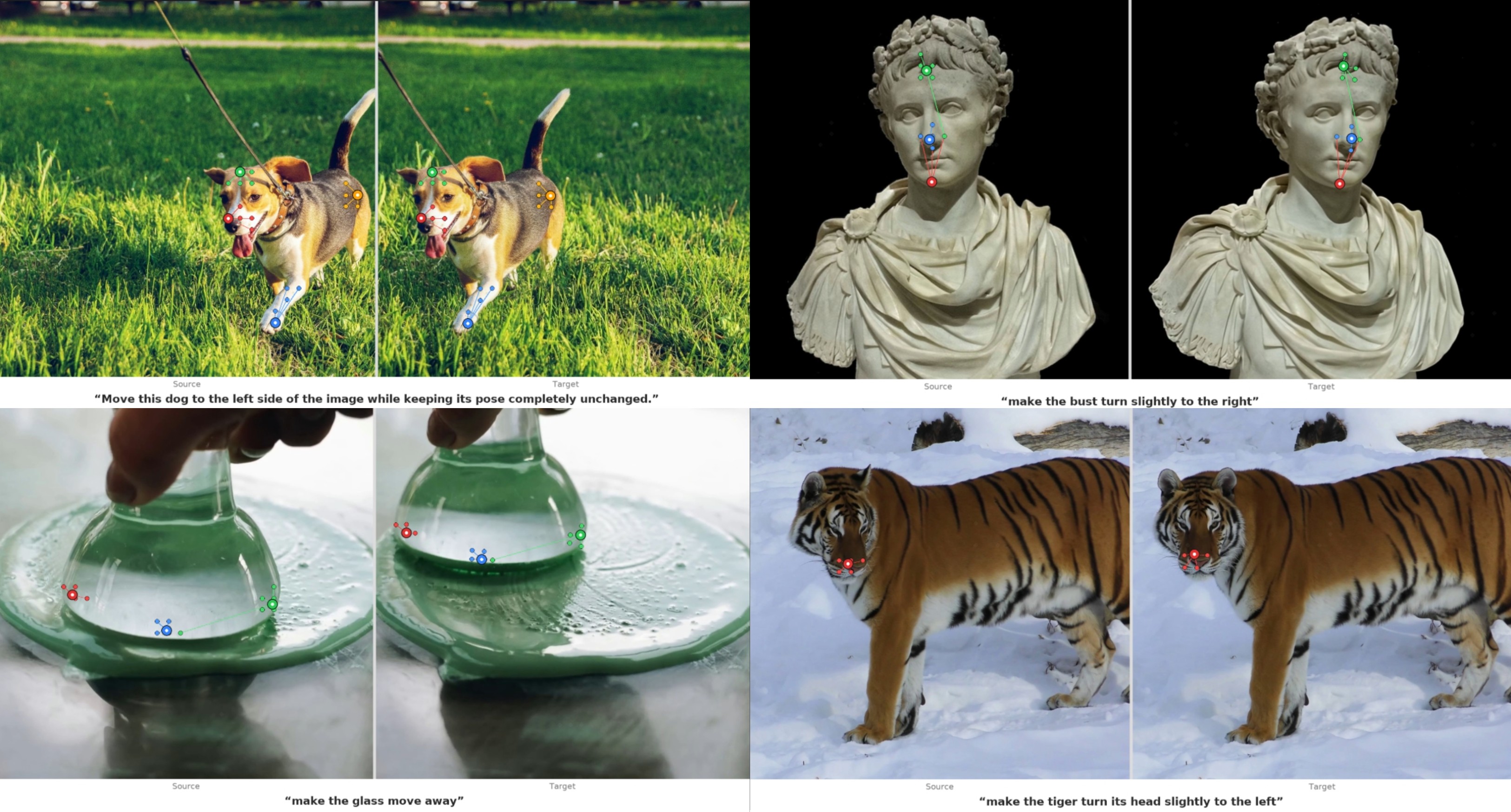}
    \caption{More samples in TV-Edit-Bench.}
    \label{fig:appen_benchshow}
\end{figure}

\subsection{Samples in TV-Edit-Bench}
As illustrated in \cref{fig:appen_benchshow}, we present additional samples from our TV-Edit-Bench dataset. Each pair consists of a source image (left) and a reference target image (right), which maintain high visual consistency owing to our rigorous generation pipeline and manual filtering. The annotated points indicate spatial correspondences between the two images. Specifically, the larger points serve as visual prompts for the editing process and are utilized to compute the sparse MD (\(\mathrm{MD_s}\)) during evaluation. Recognizing that a natural spatial edit typically involves the holistic transformation of a local region, we sample additional surrounding points to track and evaluate the dense MD (\(\mathrm{MD_d}\)) between the edited results and the targets.

\begin{table}[t]
  \centering
  \caption{Comparison of drag-based editing datasets.}
  \label{tab:benchmark}
  \setlength{\tabcolsep}{4pt}
  \renewcommand\theadfont{\bfseries}
  \begin{tabular}{lcccccc}
    \toprule
    \thead{Dataset} & \thead{\#Train/\\\#Test} & \thead{Target\\Image} & \thead{Instruction} & \thead{Description} & \thead{Dense\\Annotation} & \thead{Controlled\\Variation} \\
    \midrule
    DragBench \cite{mou2023dragondiffusion}     & 0 / 205      & \xmark & \xmark & \cmark & \xmark & \xmark \\
    LightningDrag \cite{shi2024lightningdrag} & 220k / 0     & \cmark & \xmark & \cmark & \xmark & \xmark \\
    FramePainter \cite{zhang2025framepainter}  & 22k / 200      & \cmark & \xmark & \cmark & \xmark & \xmark \\
    RealDrag \cite{zafarani2025realdrag}      & 0 / 415      & \cmark & \cmark & \cmark & \xmark & \xmark \\
    \midrule
    \textbf{Ours} & {23k / 120} & \cmark & \cmark & \cmark & \cmark & \cmark \\
    \bottomrule
  \end{tabular}
\end{table}

Furthermore, \cref{tab:benchmark} compares our established dataset with existing drag-based datasets, highlighting our unique advantages. Our dataset provides comprehensive splits for both training and evaluation, with each image pair accompanied by a specific motion-centric instruction. Notably, TV-Edit-Bench is the first to introduce dense annotations, which are crucial for evaluating the consistency of regional transformations. Additionally, we incorporate test pairs with controlled variations, including both magnitude and semantic variation cases, to comprehensively assess the spatial accuracy and semantic flexibility of various editing methods.

\subsection{Evaluation Protocol}

\textbf{Details of DINOv3-based evaluation metrics.}
We evaluate editing quality using features from DINOv3 ~\cite{simeoni2025dinov3}.
We denote by $f_{\mathrm{cls}}(I)$ and $\{f_i^{\mathrm{patch}}(I)\}_{i=1}^{N}$ the
CLS token and the set of $N$ patch tokens extracted from image $I$, respectively.

\textit{- Global DINO Score.}
The overall semantic consistency between the target image $I_{\mathrm{tgt}}$
and the edited image $I_{\mathrm{edit}}$ is measured via the cosine similarity of their CLS tokens:
\begin{equation}
  DS_{\mathrm{global}}
  = \cos\!\bigl(f_{\mathrm{cls}}(I_{\mathrm{tgt}}),\;
                 f_{\mathrm{cls}}(I_{\mathrm{edit}})\bigr).
  \label{eq:global_dino}
\end{equation}

\textit{- Local DINO Score.}
To assess the fine-grained fidelity within edited regions, we employ a nearest-neighbour
patch matching scheme. Given a reference image $I_{\mathrm{ref}}$ (either source or
target) and a corresponding binary mask $M$, we first downsample $M$ to the patch
grid via average pooling to obtain the set of masked
patch indices $\mathcal{M}$. For each masked reference patch $i\!\in\!\mathcal{M}$,
we find its nearest neighbour in the edited image:
\begin{equation}
  j^{*}(i)
  = \argmax_{j \in \{1,\dots,N\}}
    \cos\!\bigl(\hat{f}_i^{\mathrm{patch}}(I_{\mathrm{ref}}),\;
                \hat{f}_j^{\mathrm{patch}}(I_{\mathrm{edit}})\bigr),
  \label{eq:local_nn}
\end{equation}
where $\hat{f} = f / \|f\|$ means $\ell_2$-normalization. The local score is the
average similarity over all masked patches:
\begin{equation}
  DS_{\mathrm{local}}
  = \frac{1}{|\mathcal{M}|}
    \sum_{i \in \mathcal{M}}
    \cos\!\bigl(\hat{f}_i^{\mathrm{patch}}(I_{\mathrm{ref}}),\;
                \hat{f}_{j^{*}(i)}^{\mathrm{patch}}(I_{\mathrm{edit}})\bigr).
  \label{eq:local_dino}
\end{equation}
By computing similarity at the patch level, this approach effectively decouples content fidelity from spatial positioning. Consequently, it mitigates the evaluation inaccuracies that arise when using strictly pixel-aligned metrics such as LPIPS \cite{lpips}.

\textbf{Details of MLLM-based evaluation metrics.}
To comprehensively evaluate our model, we adopt the automatic evaluation protocol introduced in ContextDrag \cite{he2025contextdrag} based on DreamBench++ \cite{peng2024dreambench++}. Specifically, we leverage an MLLM (Qwen-3-VL \cite{bai2025qwen3}) as the evaluator to assess the generated images across two dimensions: prompt following (PF) and concept preservation (CP).
We modify the original prompts from ContextDrag by incorporating reference ground truth from our benchmark dataset as an evaluation criterion. This addition provides clearer guidance, making it easier for the MLLM to understand the evaluation task.

\begin{figure}[H]
    \centering
    \includegraphics[width=\linewidth]{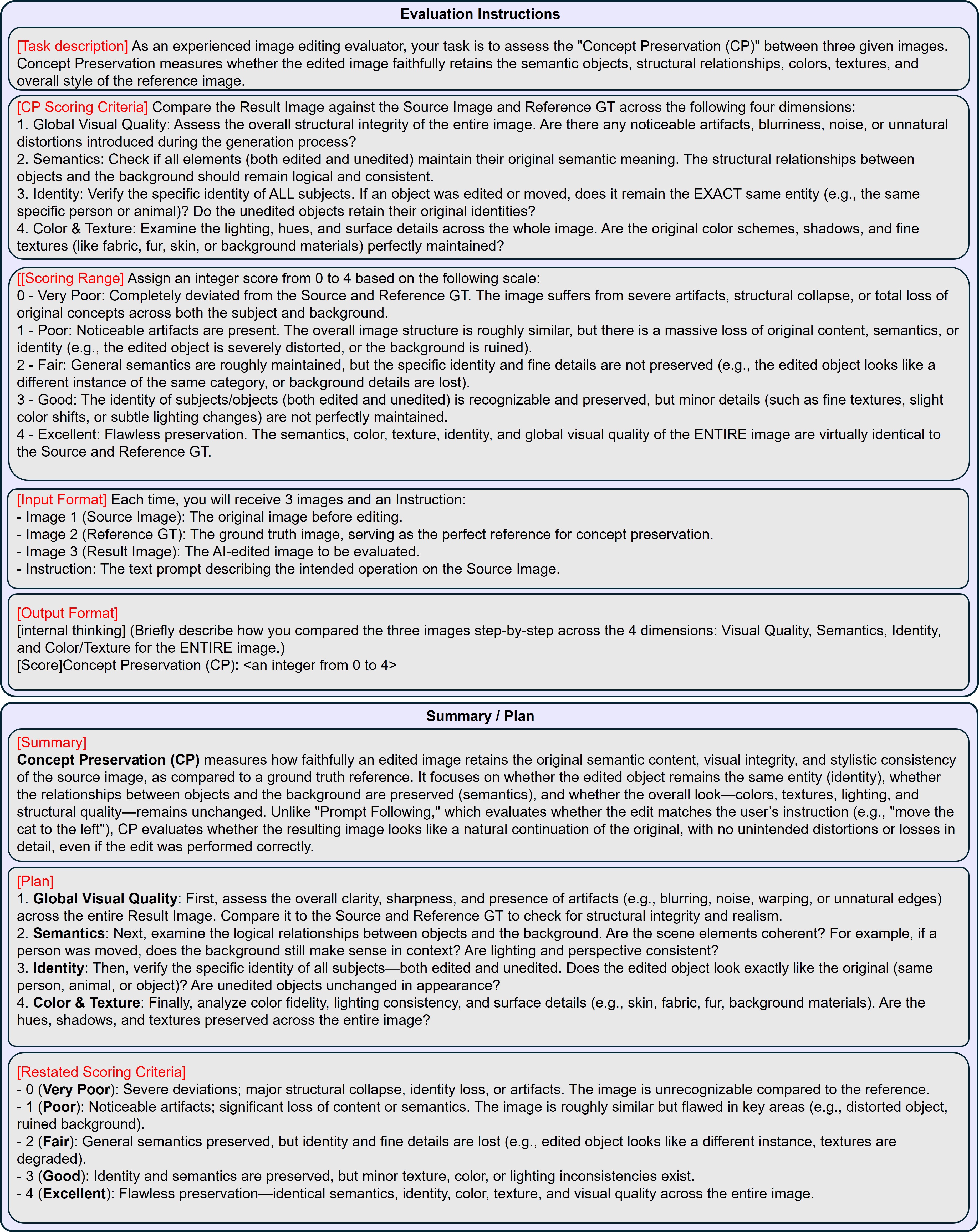}
    \caption{Illustration of the Concept Preservation (CP) Evaluation Instruction and the corresponding Summary / Planning returned by Qwen 3-VL.}
    \label{fig:app_cp}
\end{figure}

\textit{- Concept Preservation.}
CP employs an MLLM for a comprehensive evaluation of image fidelity, surpassing traditional pixel- or feature-level metrics by assessing the semantic consistency of unedited regions. \cref{fig:app_cp} shows the evaluation instructions for MLLM. Prompted with the \textbf{source image}, \textbf{reference GT}, \textbf{edited result}, and \textbf{textual instruction}, the MLLM utilizes self-aligned reasoning to understand and plan the evaluation process. Finally, it assigns a score from 0 to 4 for each case. The final score for a method is obtained by averaging and normalizing these case-level ratings.

\begin{figure}[H]
    \centering
    \includegraphics[width=\linewidth]{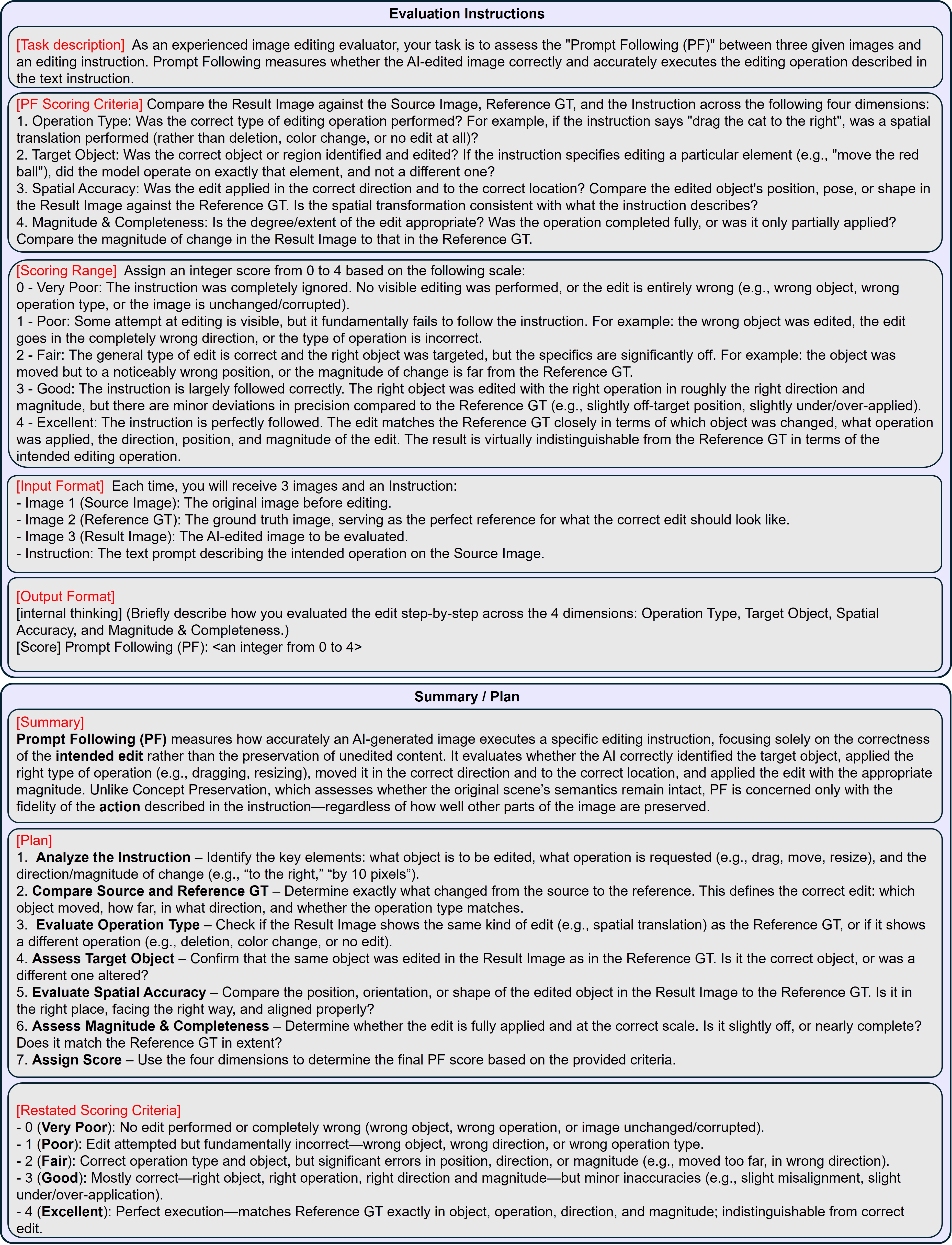}
    \caption{Illustration of the Prompt Following (PF) Evaluation Instruction and the corresponding Summary / Planning returned by Qwen 3-VL.}
    \label{fig:app_pf}
\end{figure}

\textit{- Prompt Following.} 
PF evaluates whether the edited result adheres to the user-provided textual instruction and successfully achieves the intended semantic transformation. Compared to the evaluation prompts used in ContextDrag \cite{he2025contextdrag}, our TV-Edit-Bench introduces a key improvement: rather than requiring the MLLM to infer semantic changes from visual prompts, we directly supply the explicit semantic transformations. This modification yields significantly more reliable evaluation results. As illustrated in \cref{fig:app_pf}, we provide the MLLM with the source image, reference ground truth, edited result, and textual instruction, prompting it to rate the editing correctness ranging from 0 to 4. Specifically, the MLLM is instructed to focus on four key aspects, including operation type, target object, spatial accuracy, and magnitude. After self-aligned reasoning, the model formulates a 6-step evaluation plan to assign the score. Finally, similar to the previous metric, the case-level scores are averaged and normalized to produce the final performance score.

\section{More Editing Results of TV-Edit}
\subsection{Visual Comparisons on TV-Edit-Bench}
\begin{figure}[H]
    \centering
    \includegraphics[width=\linewidth]{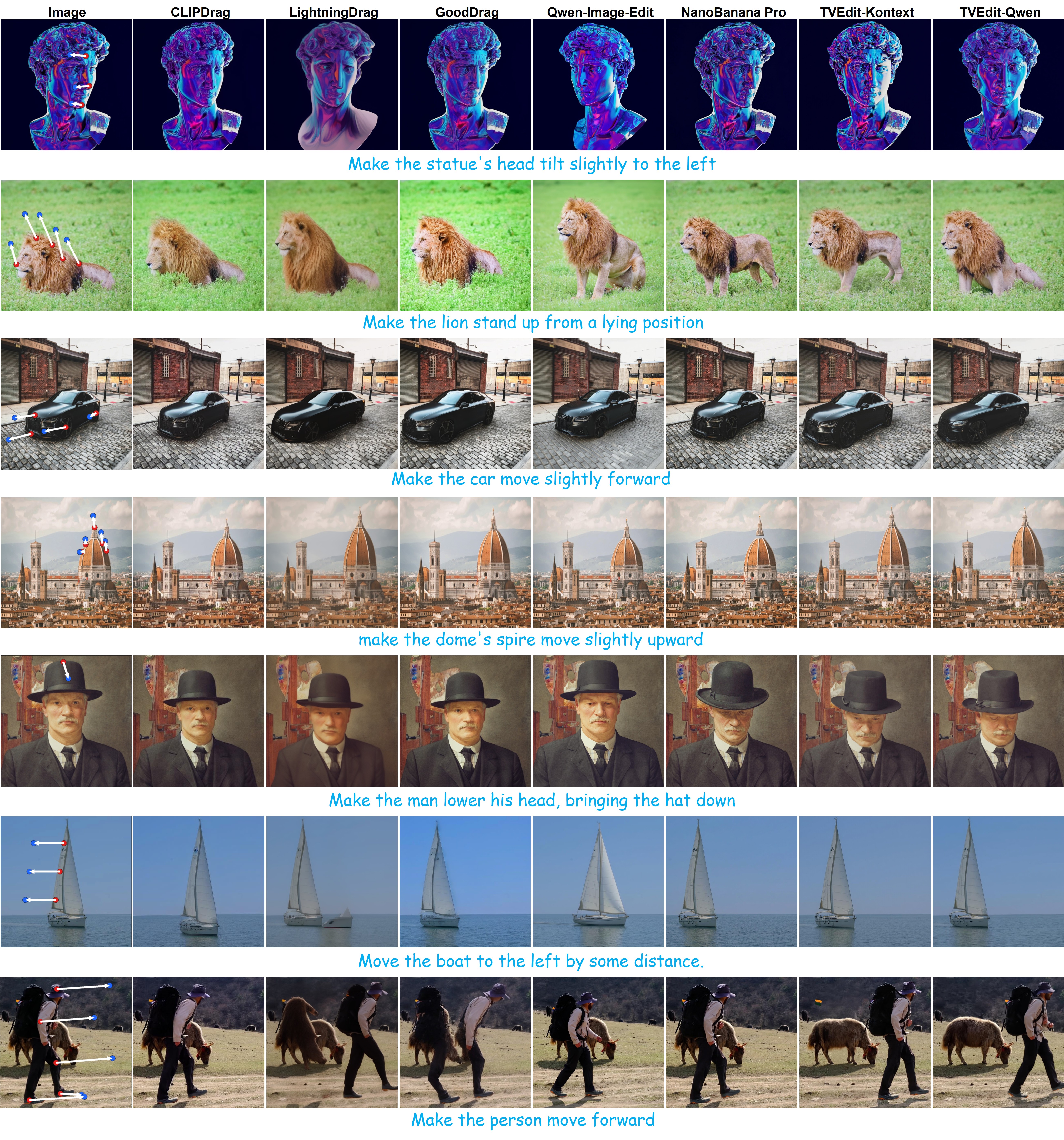}
    \caption{Visual comparison on TV-Edit-Bench. Best viewed by zoom-in.}
    \label{fig:appen_viscomp}
\end{figure}

As shown in \cref{fig:appen_viscomp}, we provide more visual comparisons on TV-Edit-Bench. 
TV-Edit is capable of performing a variety of spatial transformations, including rotation and translation, demonstrating superior image fidelity and geometric accuracy compared to other methods. Furthermore, it is evident that TV-Edit is highly robust to edits across a wide range of magnitudes and affected areas. For instance, as shown in the last row, moving the person to the right side of the image involves a substantially large transformation magnitude and spatial extent. Despite this, TV-Edit still yields high-quality editing results, whereas other baseline methods suffer from severe artifacts.

\subsection{Results on Simultaneous Spatial Control and Semantic Editing}

\begin{figure}[H]
    \centering
    \includegraphics[width=1\linewidth]{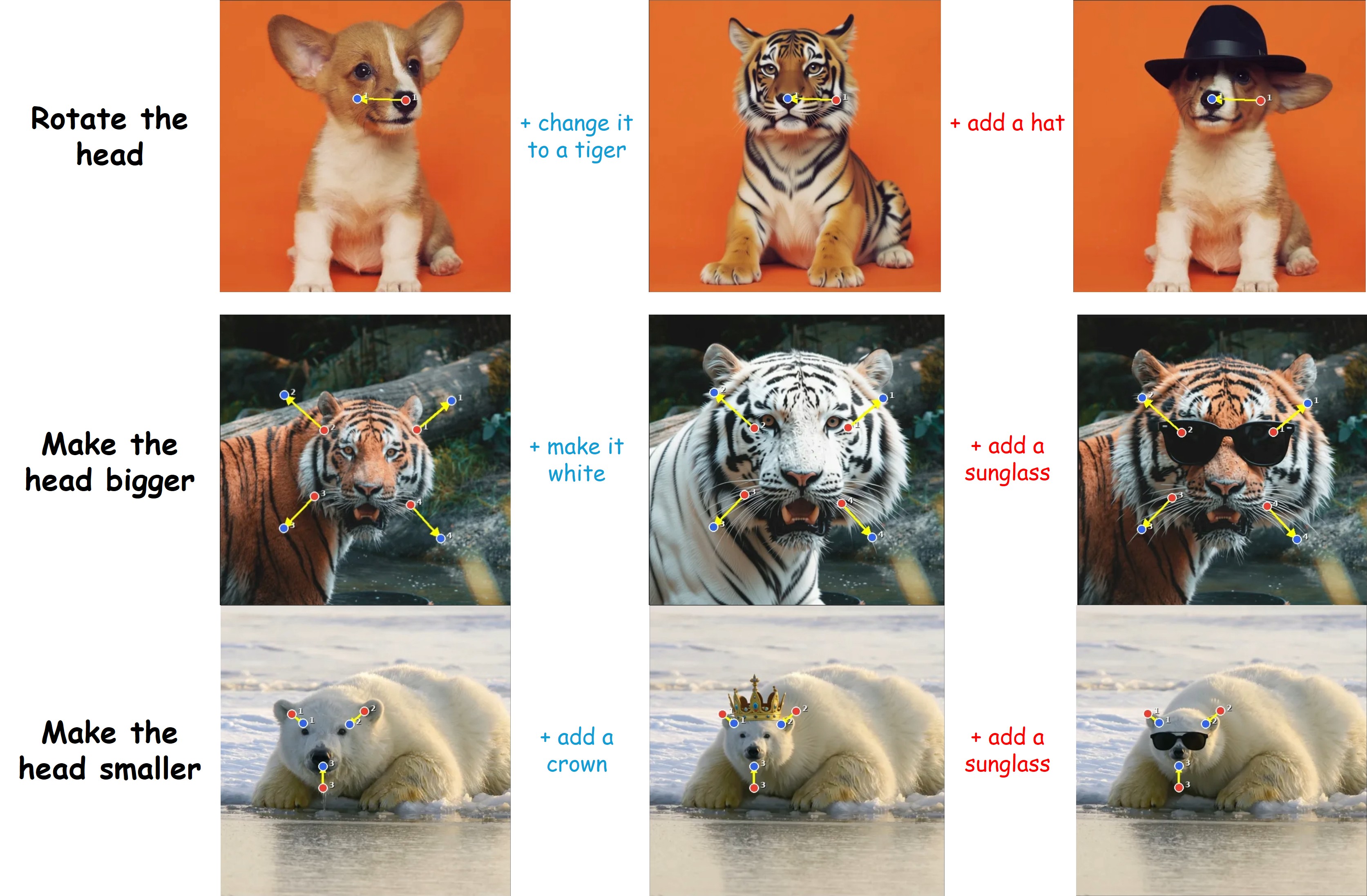}
    \caption{Results on simultaneous spatial control and semantic editing.}
    \label{fig:app_sem_edit}
\end{figure}

Our method is capable of achieving precise spatial control while simultaneously accommodating additional semantic edits. As illustrated in Figure~\ref{fig:app_sem_edit}, by providing the instruction ``change it to a tiger'' alongside the motion constraints, our model successfully rotates the dog's head while transforming its identity into a tiger. Similarly, as shown in the second and third rows, our approach allows for complex semantic modifications—such as adding objects or altering colors—while maintaining precise control over the magnitude of spatial transformations.

\section{More Ablation Studies on TV-Edit}

{\textbf{Ablation of textual and visual inputs at inference time.}}
\begin{table}[t]
\centering
\caption{Inference-time ablation of textual and visual inputs.}
\label{tab:ablation5}
\begin{tabular}{llcc}
\hline
Exp. & Setting & MD$_d \downarrow$ & PF $\uparrow$ \\
\hline
1 & Qwen baseline & .1380 & .86 \\
2 & Trajectory + text (w/o controller text) & .0502 & .91 \\
3 & Trajectory only (empty text) & .0627 & .83 \\
4 & Text only (w/o trajectory) & .1410 & .87 \\
5 & Trajectory + text (full model) & \textbf{.0462} & \textbf{.93} \\
\hline
\end{tabular}
\label{tab:ablation_input}
\end{table}
\cref{tab:ablation5}  presents the effect of different combinations of textual and visual inputs on semantic instruction following. Specifically, for trajectory-only inference, we remove the text input either from the control branch alone or from the entire model. Comparing Exp. 1 with Exp. 4, we see that text-only TV-Edit performs similarly to Qwen-Image-Edit baseline, indicating that spatial control is unavailable without trajectories. Comparing Exp. 2 with Exp. 5, we see that removing text from the control branch causes only minor degradation, demonstrating the robustness of the spatial branch, while the PF drop also shows that controller-side text benefits semantic execution. Comparing Exp. 3 with Exp. 5, we see that with completely empty text, the model retains reasonable geometric accuracy but suffers from a clear PF drop from 0.93 to 0.83, as it cannot distinguish actions with similar trajectories such as opening the mouth and raising the head. Finally, comparing Exp. 4 with Exp. 5, we see that adding visual prompts significantly reduces MD$_\mathrm{d}$ from 0.1410 to 0.0462 and PF from 0.87 to 0.93, showing that visual input provides spatial constraints and facilitates semantic execution through collaboration with text. Overall, text resolves semantic ambiguity, visual prompts provide spatial control, and their collaboration yields the best semantic and spatial accuracy.

\begin{table}[t]
    \centering
    \caption{Results of text-only LoRA fine-tuning on TV-Edit-23K.}
    \label{tab:ablation6}
    \begin{tabular}{l c c c c}
        \toprule
        \textbf{Method} & {DS$_\mathrm{global}^\mathrm{tar}$} $\uparrow$ & {DS$_\mathrm{local}^\mathrm{tar}$} $\uparrow$ & MD$_{d}$ $\downarrow$ & PF $\uparrow$ \\
        \midrule
        Qwen-Image-Edit & .8575 & .9191 & .1379 & .86 \\
        + text-only LoRA & .8806 & .9230 & .1254 & .87 \\
        \bottomrule
    \end{tabular}
\end{table}
{\textbf{Text-only LoRA fine-tuning on TV-Edit-23K.}}
\cref{tab:ablation6} shows the comparison of LoRA fine-tuning on Qwen-Image-Edit using instruction–image pairs from TV-Edit-23K, in order to disentangle the contributions of data and model architecture. Training on a large collection of motion-related samples better adapts the model to such editing patterns, leading to a moderate improvement on the benchmark. However, the gain remains limited, and the relatively high MD$_d$ indicates that text alone cannot reliably control motion magnitude. This suggests that text-only supervision is insufficient for modeling motion magnitude, and multimodal architecture and joint training paradigm are the keys to effectively improving performance on the TV-Edit task.

\section{Comparison on Drag-Bench}
\begin{table}[t]
  \centering
  \caption{Quantitative comparison of drag-based editing methods on DragBench.}
  \label{tab:comparison}
  \begin{tabular}{lcc}
    \toprule
    Method & IF ($\uparrow$) & MD ($\downarrow$) \\
    \midrule
    DragDiffusion  \cite{mou2023dragondiffusion}           & 0.88          & 32.13 \\
    DragNoise \cite{liu2024drag} &0.89 &35.17 \\
    AdaptiveDrag \cite{chen2024adaptivedrag}&0.86 &35.70 \\
    CLIPDrag \cite{jiang2024clipdrag}&0.88 &32.30 \\
    GoodDrag \cite{zhang2024gooddrag}         & 0.87          & 24.26 \\
    LightningDrag  \cite{shi2024lightningdrag}           & \textbf{0.89} & 29.10 \\
    DragLora  \cite{xia2025draglora} & 0.87          & 23.77 \\
    GeoDrag \cite{pu2025dragging}    & 0.85          &29.24  \\
    \midrule
    TV-Edit-Qwen (ours)             & 0.86 & \textbf{17.31} \\
    \bottomrule
  \end{tabular}
\end{table}
To demonstrate the generalization ability of our approach in terms of geometric accuracy,  we follow prior work \cite{xia2025draglora} and conduct experiments on the drag-based benchmark DragBench \cite{mou2023dragondiffusion}, comparing our method both quantitatively and qualitatively against several state-of-the-art baselines.

\subsection{Quantitative Comparison}
 For inference on DragBench, we first employ Qwen-3-VL \cite{bai2025qwen3} to analyze the images alongside their visual prompts, then utilize the inferred semantic changes as textual instructions for TV-Edit-Qwen. Following prior work \cite{xia2025draglora, mou2023dragondiffusion}, we evaluate our method using MD and IF, which are computed as the mean point displacement distance and 1$-$LPIPS, respectively.
 As shown in \cref{tab:comparison}, TV-Edit-Qwen achieves an excellent balance between geometric accuracy and image fidelity. Notably, it yields an MD of 17.31, which is significantly lower than existing methods. Such a low MD indicates a highly successful geometric edit, which inevitably introduces some deviation from the original image. Nevertheless, even without requiring pre-defined edit region masks, TV-Edit-Qwen achieves an IF score of 0.86, surpassing GeoDrag and comparable to most state-of-the-art approaches.

\subsection{Qualitative Comparison}
\begin{figure}[H]
    \centering
    \includegraphics[width=1\linewidth]{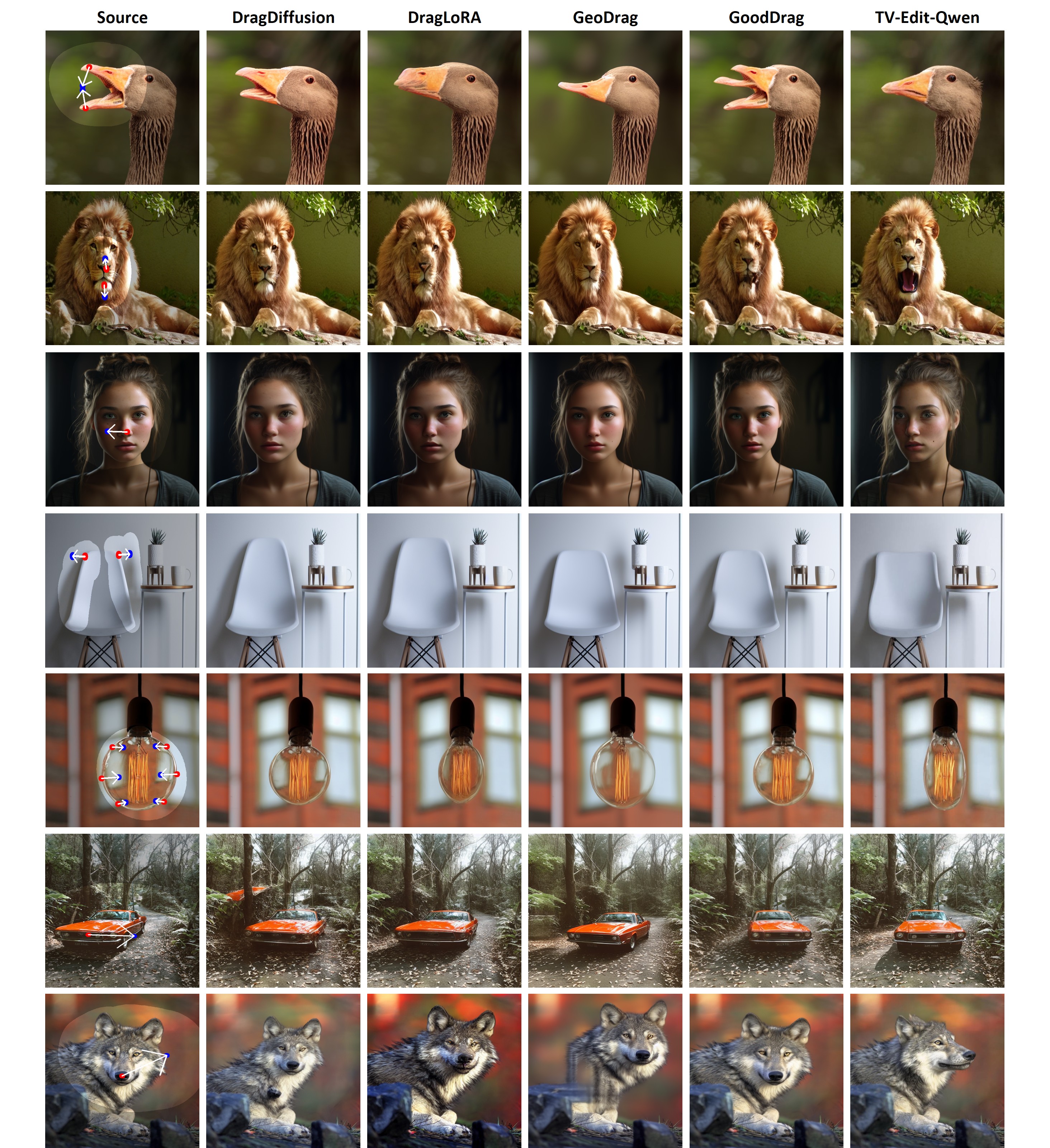}
    \caption{Visual comparisons among TV-Edit and drag-based methods on DragBench.}
    \label{fig:appen_dragbench}
\end{figure}

Figure~\ref{fig:appen_dragbench} compares TV-Edit-Qwen with state-of-the-art drag-based methods, including  DragDiffusion \cite{mou2023dragondiffusion}, DragLora \cite{xia2025draglora}, GeoDrag \cite{pu2025dragging} and GoodDrag \cite{zhang2024gooddrag}. From the visual comparisons, one can see that our TV-Edit-Qwen achieves much better geometric accuracy and fidelity. 
Specifically, in the second row where the visual prompt aims to open the lion's mouth, only TV-Edit-Qwen successfully executes this semantic transformation. GoodDrag attempts to follow the prompt but introduces artifacts, while the other methods show no noticeable changes. This highlights the inherent difficulty of previous visual-prompt-only methods in handling complex semantic actions. Furthermore, the sixth row demonstrates our model's robust spatial editing capabilities: TV-Edit-Qwen correctly rotates the car head with high visual quality. In contrast, among the baselines, only GoodDrag exhibits a visible rotation, whereas DragDiffusion fails to achieve the edit and generates artifacts.

\begin{figure}[H]
    \centering
    \includegraphics[width=1\linewidth]{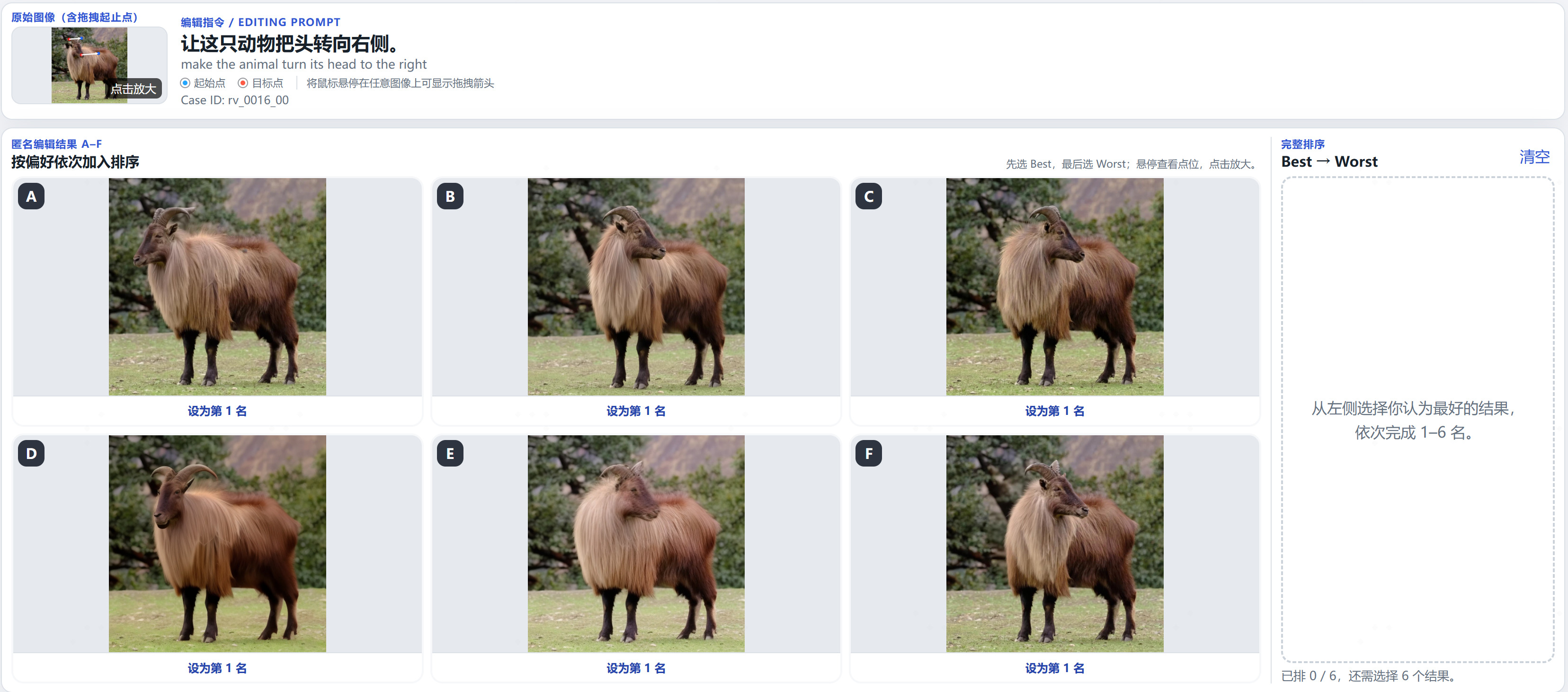}
    \caption{User study web interface.}
    \label{fig:appen_userstudy}
\end{figure}

\begin{figure}[H]
    \centering
    \includegraphics[width=\linewidth]{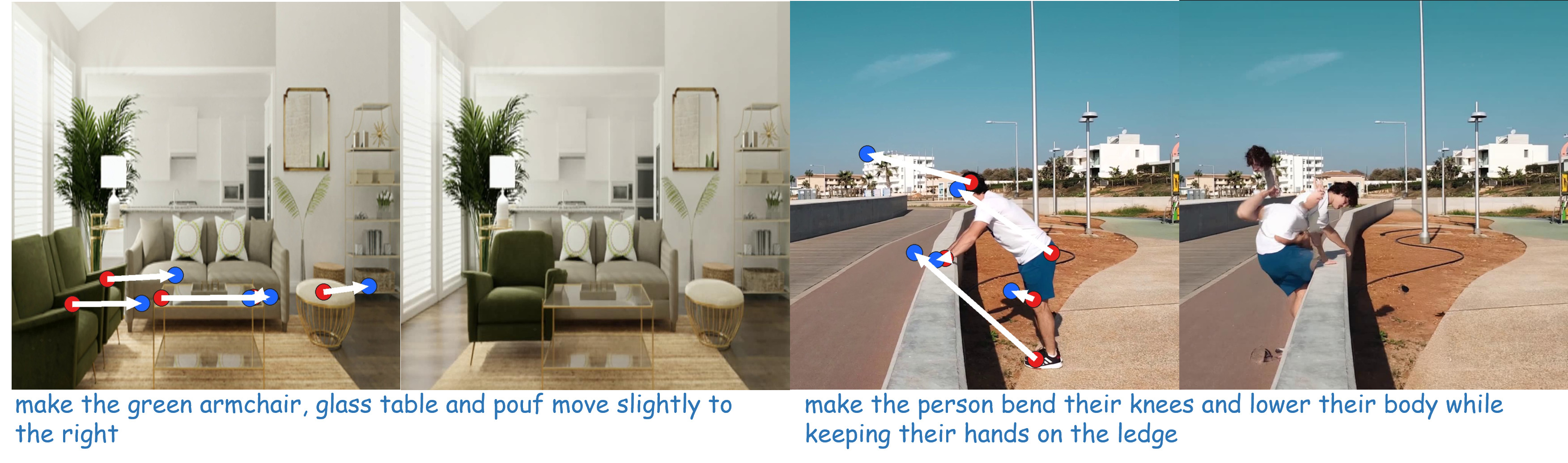}
    \caption{Failure cases.}
    \label{fig:appen_failure}
\end{figure}

\section{User Study}
\begin{table}[t]
    \centering
    \caption{User study on overall editing quality.}
    \label{tab:user_study_overall}
    \begin{tabular}{lcccccc}
        \toprule
        \textbf{Metric} & \textbf{TV-Qwen} & \textbf{TV-Kontext} & \textbf{NanoBanana} & \textbf{GoodDrag} & \textbf{LongCat} & \textbf{LightningDrag} \\
        \midrule
        Avg. Rank $\downarrow$ & \textbf{2.4905} & 2.5048 & 3.5439 & 3.7476 & 4.0429 & 4.6714 \\
        Top-1 (\%) $\uparrow$ & 27.62 & \textbf{30.95} & 14.76 & 8.10 & 13.33 & 5.24 \\
        Top-2 (\%) $\uparrow$ & \textbf{61.43} & 56.19 & 34.29 & 18.57 & 20.00 & 9.52 \\
        \bottomrule
    \end{tabular}
\end{table}
To further validate the effectiveness of TV-Edit, we conduct a user study comparing it against several representative methods. We select two variants of TV-Edit along with other competitive baselines that rank high in quantitative results, including NanoBanana Pro \cite{nanobanana2025}, GoodDrag \cite{zhang2024gooddrag}, LongCat-Image-Edit \cite{team2025longcat}, and LightningDrag \cite{shi2024lightningdrag}. For the experimental setup, we randomly sample 30 images from TV-Edit-Bench and anonymously present the editing results of all methods to participants via a web interface, asking them to rank the methods by overall editing quality, considering both instruction following and geometric accuracy. The user study interface is shown in \cref{fig:appen_userstudy}. A total of 15 volunteers participate, and we report the average rank of each method in \cref{tab:user_study_overall}. As shown in the table, TV-Edit-Kontext achieves the highest Top-1 rate, while TV-Edit-Qwen achieves the best average rank and Top-2 rate. Both are preferred substantially more often than instruction- and drag-based baselines, supporting our automated results.

\section{Failure Cases}
We show several failure cases of TV-Edit in \cref{fig:appen_failure}. In the left case, when multiple objects are moved simultaneously in a complex scene, the editing of some objects may fail. A possible solution is to perform multi-round editing on individual objects, which may yield better results. In the right case, the instruction mentions changes in the knees and hands but does not mention the jumping motion, which conflicts with the trajectory condition to some extent. Under such complex motion cases, the editing collapses and generates artifacts. For controlling complex motions, more accurate instruction descriptions and consistent trajectories can help improve the editing quality.

\section{Broader Impacts}
Our proposed image editing method has several potential societal impacts, both positive and negative.
On the positive side, this method can enhance creative industries by providing artists and designers with powerful tools for content creation and modification. 
However, there are potential negative impacts to consider. The technology could be misused for creating misleading or harmful content, which could have significant implications for privacy and security.
To mitigate these risks, we suggest implementing mechanisms for monitoring and controlling the use of the technology, such as gated releases and developing tools to detect and counteract malicious uses.




\end{document}